%% file: 00main.tex
\documentclass[journal]{IEEEtran}
\IEEEoverridecommandlockouts

\usepackage{tikz}
\usepackage{comment}
\usepackage{amsmath,amssymb} 
\usepackage{color}
\usepackage{amsmath}
\usepackage{amssymb}
\usepackage{booktabs}
\usepackage{tabularx}
\usepackage{changepage}
\DeclareMathOperator{\img}{img}
\DeclareMathOperator{\SSIM}{SSIM}
\usepackage{makecell}
\usepackage{multirow}
\usepackage{afterpage}
\usepackage{xcolor}
\usepackage{float}
\usepackage{subcaption}
\usepackage[shortlabels]{enumitem}

\usepackage[breaklinks,colorlinks]{hyperref}

\newcommand\tstrut{\rule{0pt}{2.4ex}}
\newcommand\bstrut{\rule[-1.0ex]{0pt}{0pt}}

\makeatletter
\renewcommand*\env@matrix[1][\arraystretch]{%
  \edef\arraystretch{#1}%
  \hskip -\arraycolsep
  \let\@ifnextchar\new@ifnextchar
  \array{*\c@MaxMatrixCols c}}
\DeclareRobustCommand\onedot{\futurelet\@let@token\@onedot}
\def\@onedot{\ifx\@let@token.\else.\null\fi\xspace}
\makeatother

\usepackage[capitalize]{cleveref}
\crefname{section}{Sec.}{Secs.}
\Crefname{section}{Section}{Sections}
\Crefname{table}{Table}{Tables}
\crefname{table}{Tab.}{Tabs.}

\usepackage{authblk}
\makeatletter
\renewcommand\maketitle{\AB@maketitle} 
\renewcommand\AB@affilsepx{\quad\protect\Affilfont} 
\makeatother

\newcommand{\deqing}[1]{{\color{orange} [Deqing: #1]}}
\newcommand{\etal}{\emph{et al.}}

\usepackage[accsupp]{axessibility}  

\def\BibTeX{{\rm B\kern-.05em{\sc i\kern-.025em b}\kern-.08em
    T\kern-.1667em\lower.7ex\hbox{E}\kern-.125emX}}

\begin{document}

\title{Topological Regularization for Dense Prediction}


\author[1]{Deqing Fu}
\author[1]{Bradley J. Nelson}
\affil[1]{Department of Statistics, University of Chicago \authorcr \{\tt \small deqing, bradnelson\}@uchicago.edu}
\maketitle
\pagenumbering{gobble}

\begin{abstract}
    Dense prediction tasks such as depth perception and semantic segmentation are important applications in computer vision that have a concrete topological description in terms of partitioning an image into connected components or estimating a function with a small number of local extrema corresponding to objects in the image.  
    We develop a form of topological regularization based on persistent homology that can be used in dense prediction tasks with these topological descriptions.  
    Experimental results show that the output topology can also appear in the internal activations of trained neural networks which allows for a novel use of topological regularization to the internal states of neural networks during training, reducing the computational cost of the regularization. 
    We demonstrate that this topological regularization of internal activations leads to improved convergence and test benchmarks on several problems and architectures.
\end{abstract}

\begin{IEEEkeywords}
Computational topology, Topological data analysis, Persistent homology, Monocular depth estimation, Semantic segmentation
\end{IEEEkeywords}

\input{01Intro}

\input{02related_work}
\input{03tda}

\input{04internal_activations}

\input{05experiment}

\input{06conclusion}


\vspace{-2ex}
\bibliographystyle{IEEEtran}
\bibliography{reference}

\end{document}


\pagestyle{headings}
\mainmatter
\def\ECCVSubNumber{4215}  

\title{Topological Regularization for Dense Prediction} 

\titlerunning{ECCV-22 submission ID \ECCVSubNumber} 
\authorrunning{ECCV-22 submission ID \ECCVSubNumber} 
\author{Anonymous ECCV submission}
\institute{Paper ID \ECCVSubNumber}

\maketitle

\section{Source Codes}
An anonymized version of our codes is here: \url{https://anonymous.4open.science/r/cvpr10937/}. \\

We use a modified version of the TopologyLayer package which implements the union-find algorithm for persistent homology in dimension 0 (this can be significantly faster than the reduction algorithm).  An anonymized version is here:
\url{https://anonymous.4open.science/r/TopologyLayer-10937}. A pull request to the original repository will be made after review.

\section{Examples of Topology in Internal Activations}
\begin{figure*}[h]
    \centering
    \includegraphics[width=0.88\linewidth]{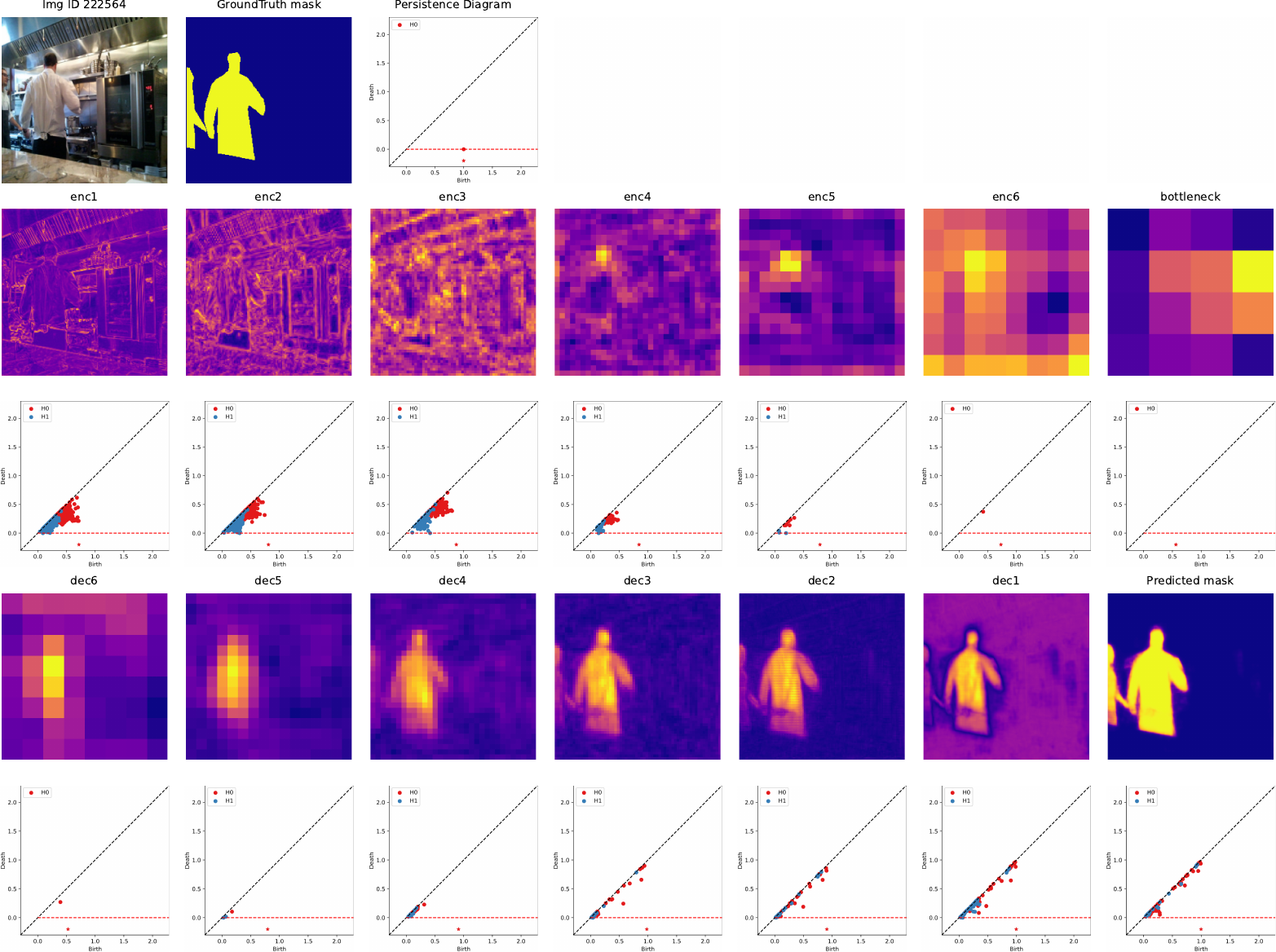}
    \caption{Example containing two human objects. Persistence diagrams remain similar after layer dec4.}
    \label{fig:example1}
\end{figure*}

\begin{figure*}[h]
    \centering
    \includegraphics[width=0.88\linewidth]{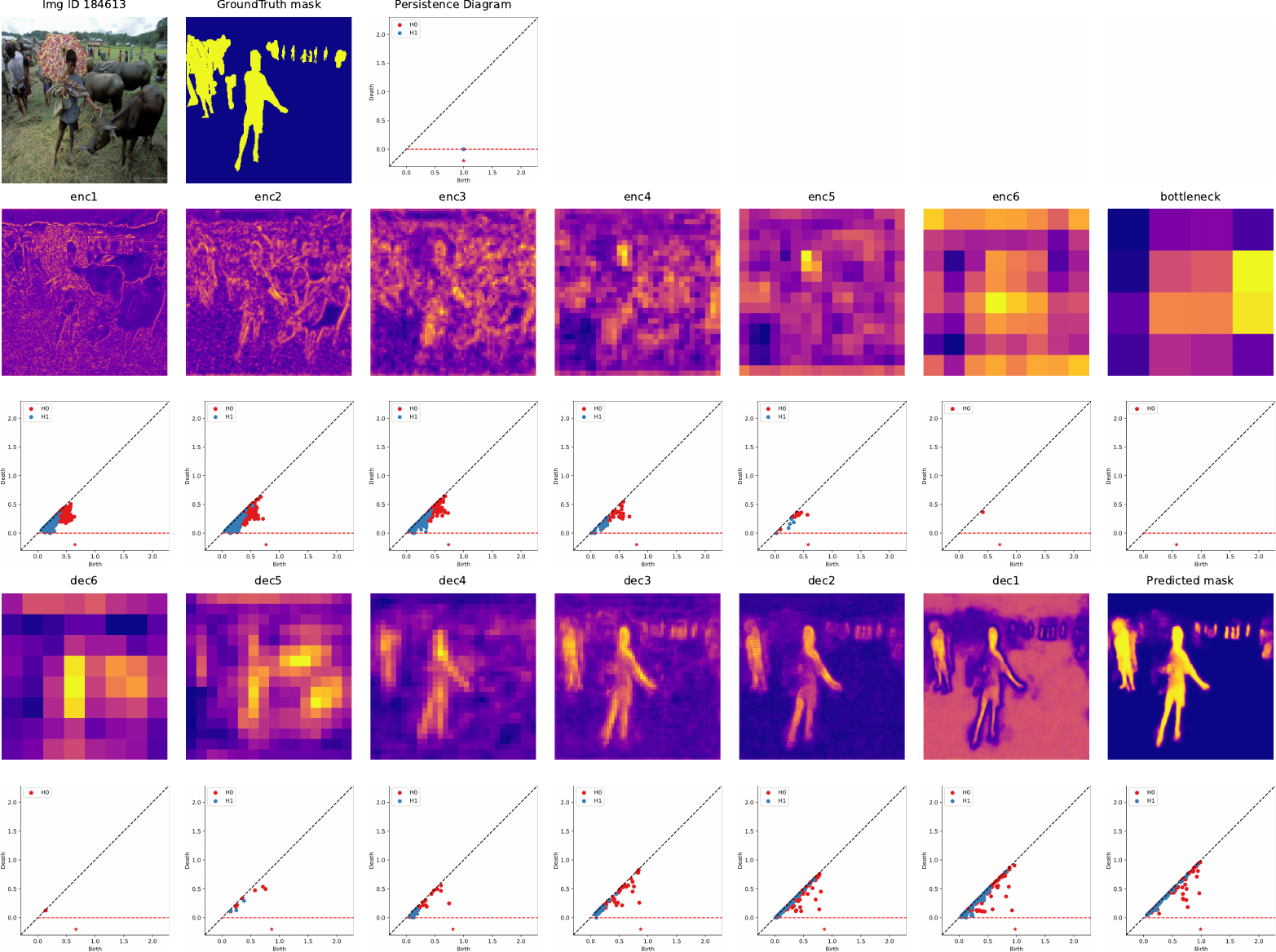}
    \caption{Example containing multiple human objects. Persistence diagrams also remain similar after layer dec4.}
    \label{fig:example2}
\end{figure*}

\begin{figure*}[h]
    \centering
    \includegraphics[width=0.88\linewidth]{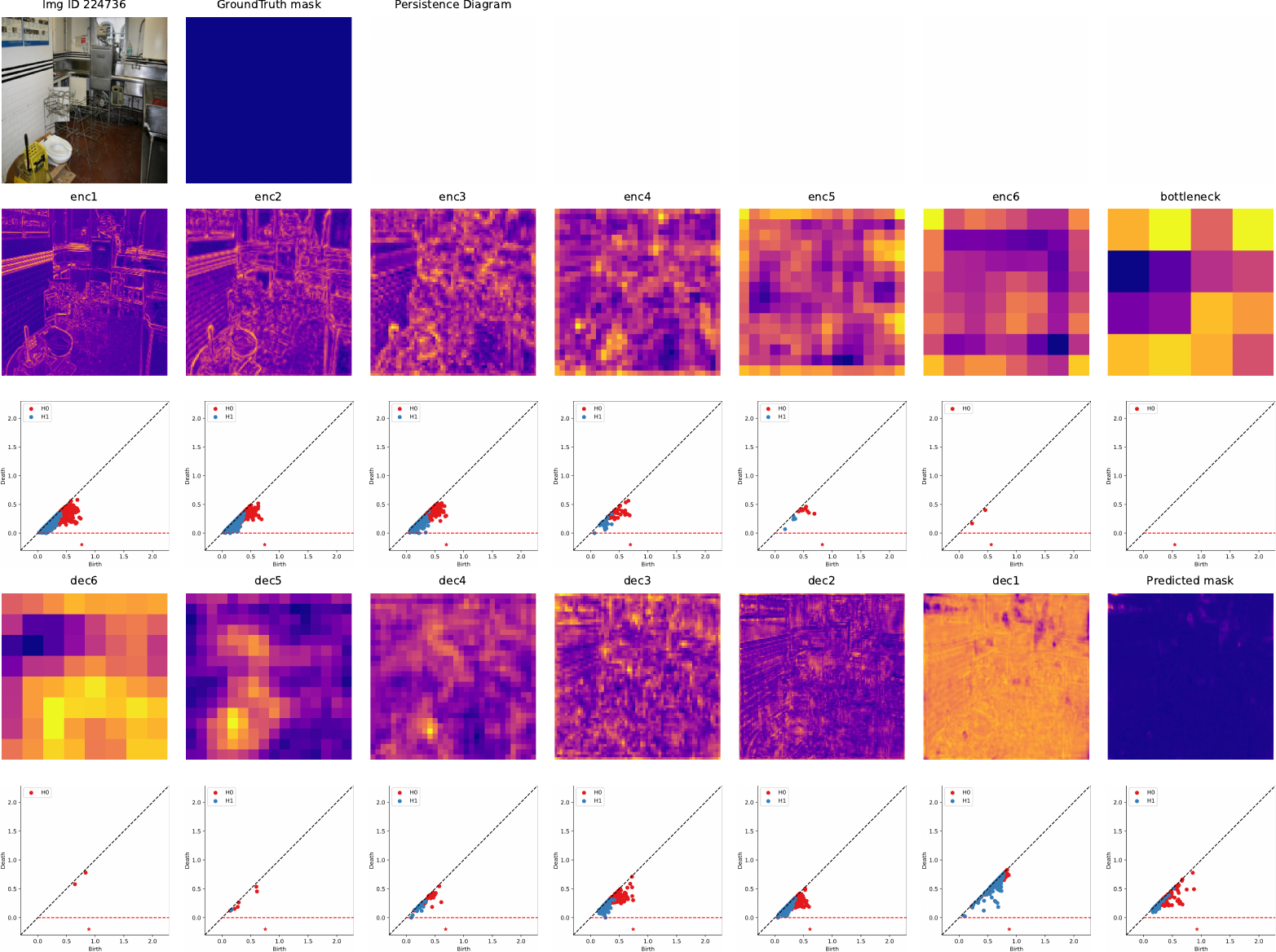}
    \caption{Example containing no human object. Persistence diagrams change during decoder layers.}
    \label{fig:example3}
\end{figure*}

\clearpage
\section{Regularization Effects of Internal Activations}
\begin{figure*}[h]
    \centering
    \begin{subfigure}[b]{0.32\linewidth}
    \includegraphics[width=\linewidth]{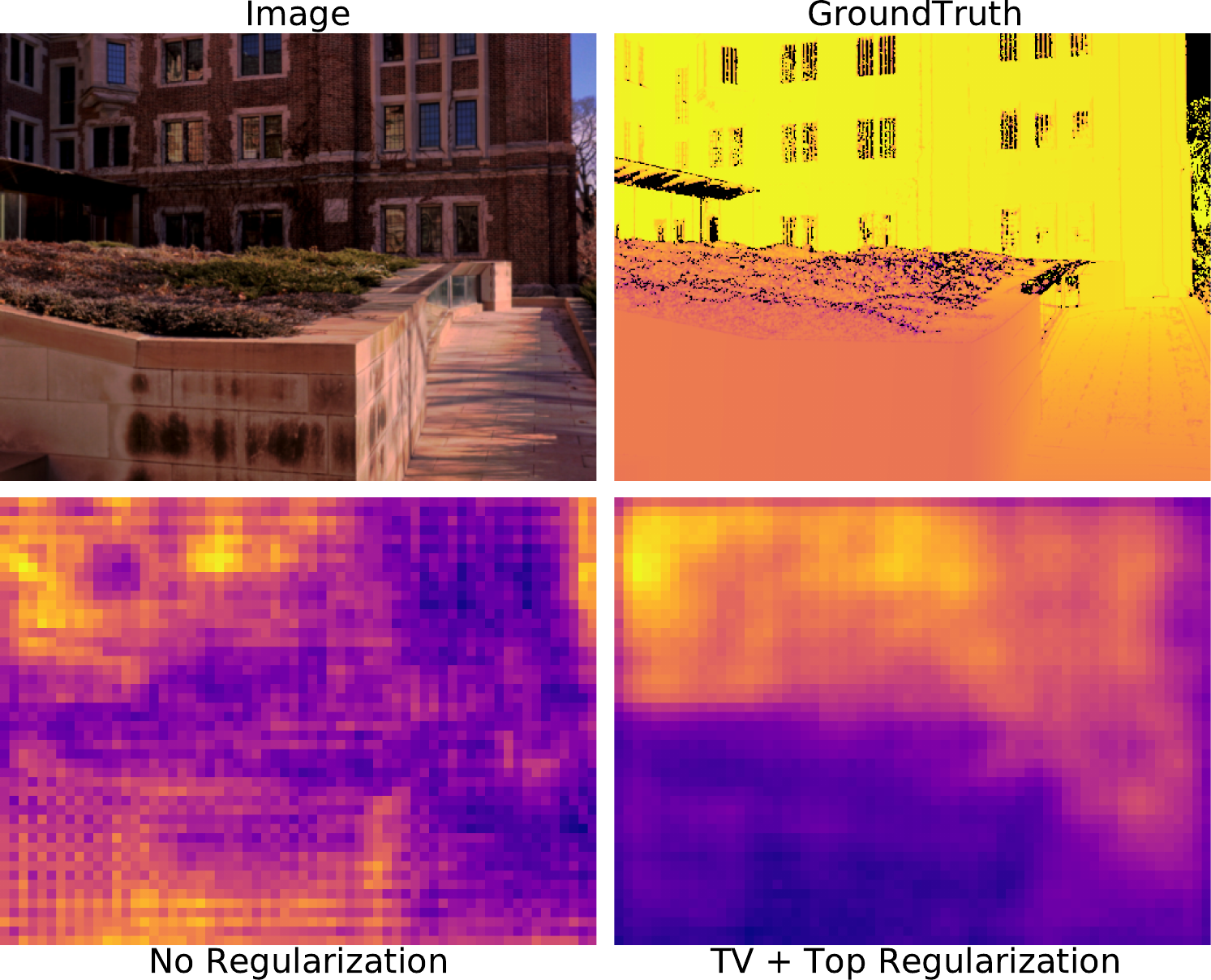}
    \end{subfigure}
    \hspace{0.2ex}
    \begin{subfigure}[b]{0.32\linewidth}
    \includegraphics[width=\linewidth]{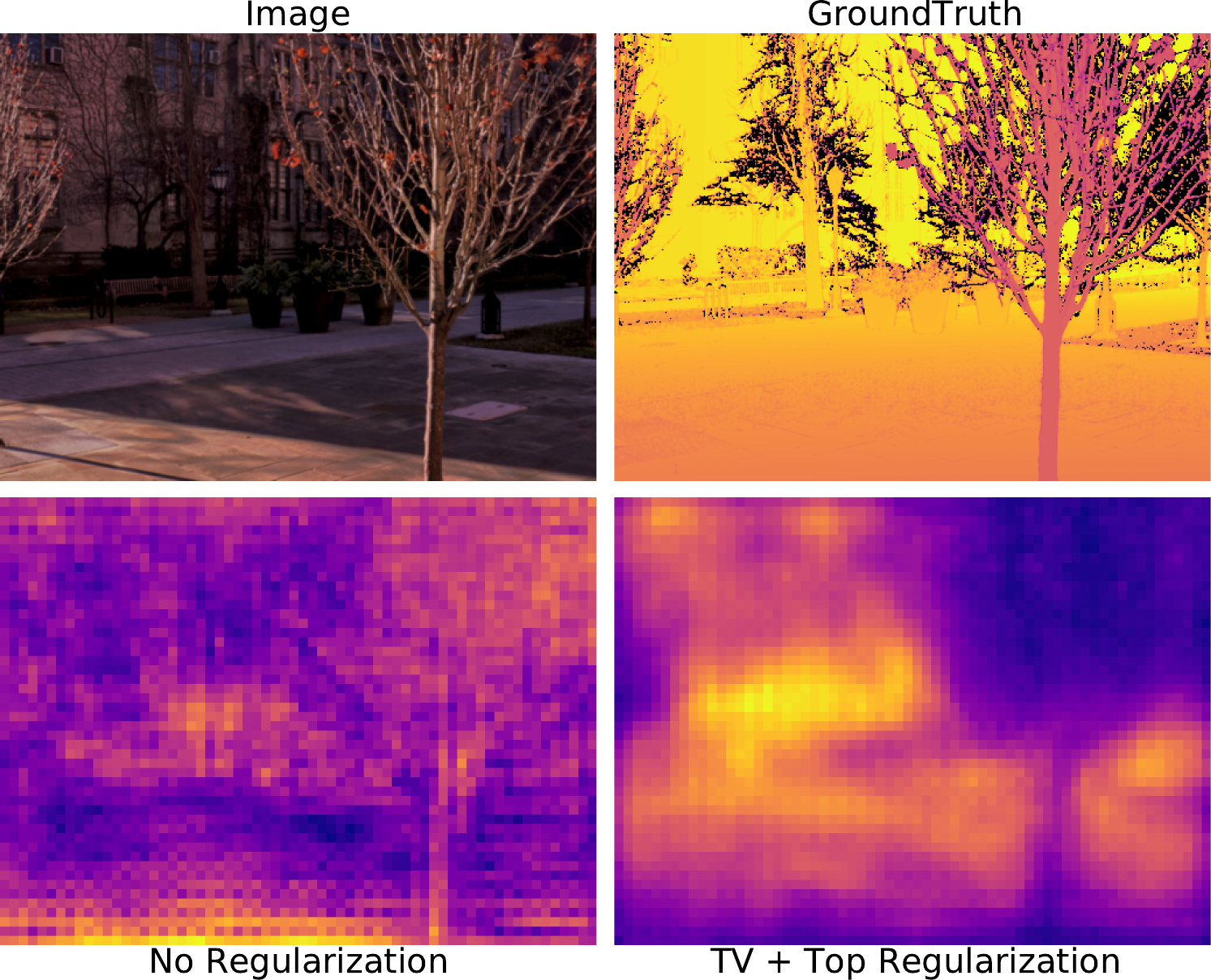}
    \end{subfigure}
    \hspace{0.2ex}
    \begin{subfigure}[b]{0.32\linewidth}
    \includegraphics[width=\linewidth]{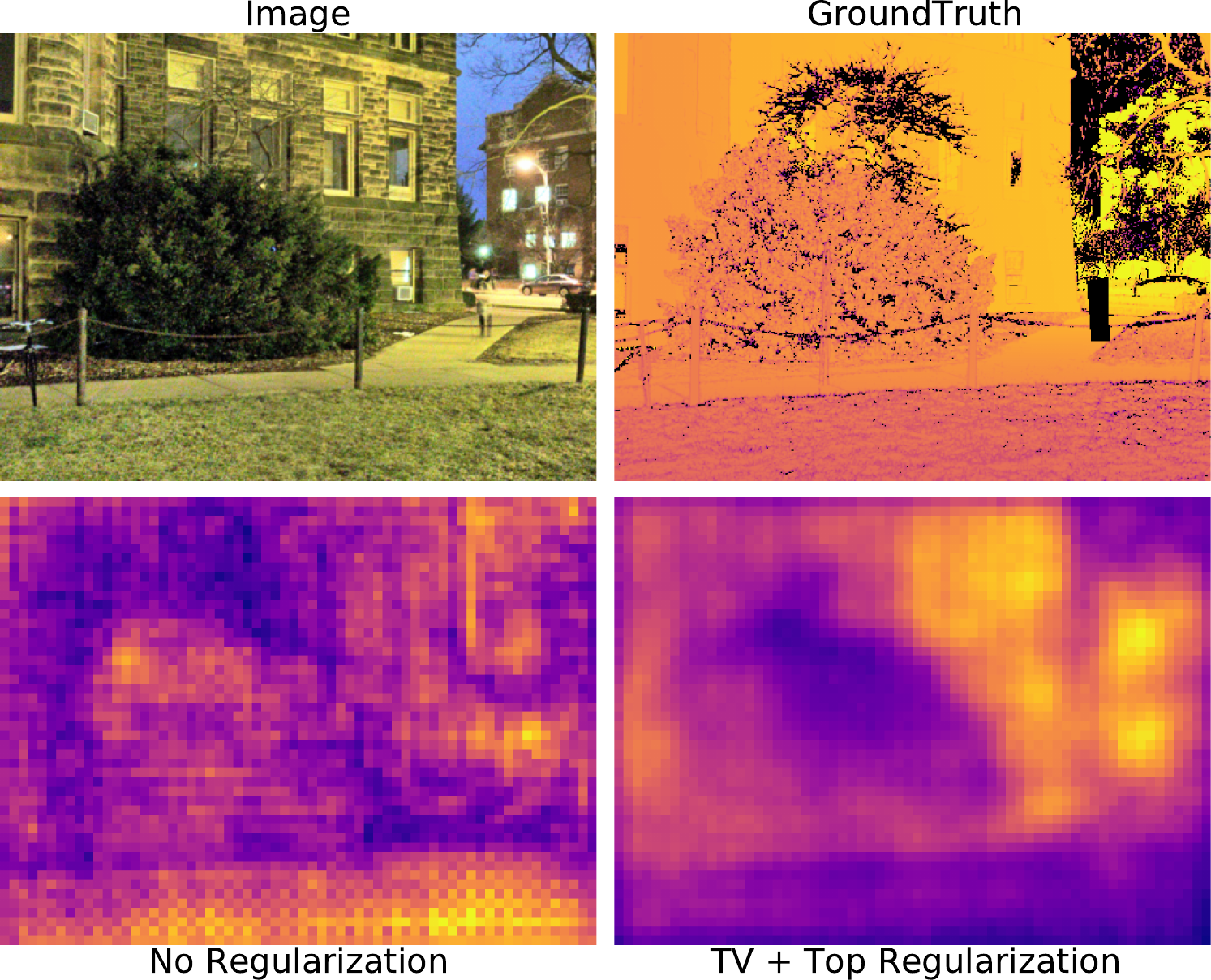}
    \end{subfigure}
    \caption{\textbf{Regularization Effects on U-Net.} The second row shows internal activations. Our regularized version helps reduce unpooling artifacts common in U-Net models. Most importantly, it keeps the interval activations concentrating on regions.}
    \label{fig:comparison_unet}
\end{figure*}

\begin{figure*}[h]
    \centering
    \begin{subfigure}[b]{0.32\linewidth}
    \includegraphics[width=\linewidth]{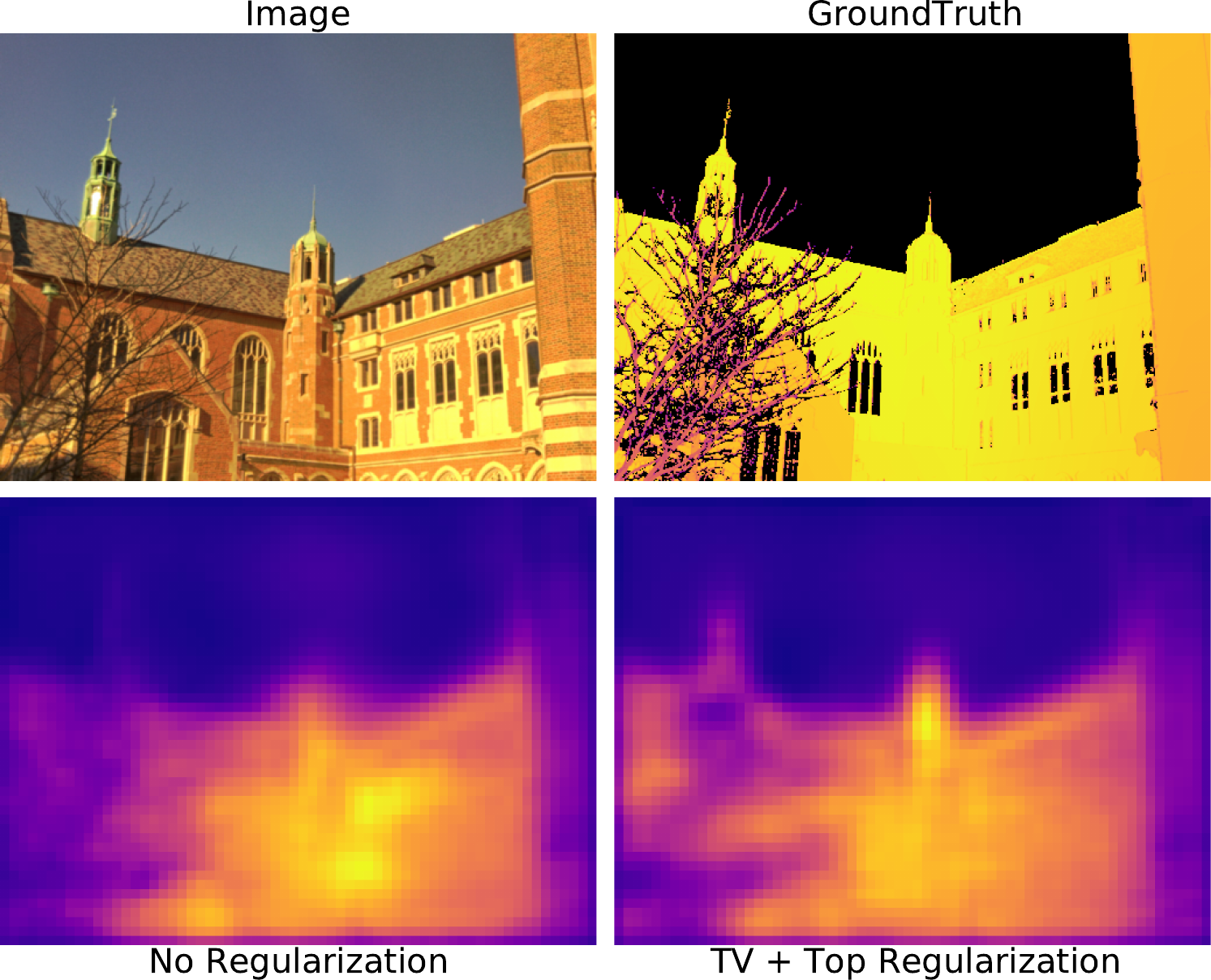}
    \end{subfigure}
    \hspace{0.2ex}
    \begin{subfigure}[b]{0.32\linewidth}
    \includegraphics[width=\linewidth]{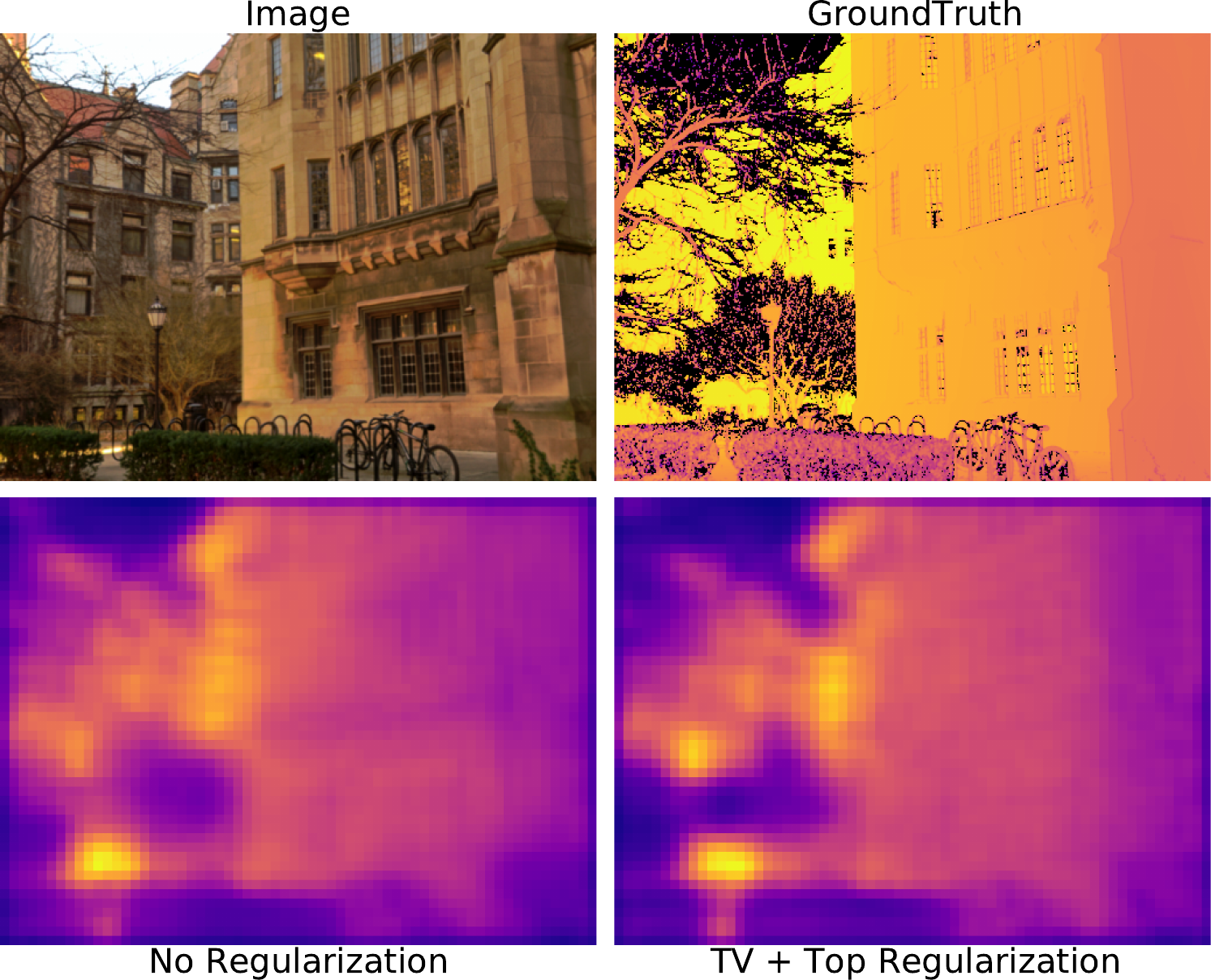}
    \end{subfigure}
    \hspace{0.2ex}
    \begin{subfigure}[b]{0.32\linewidth}
    \includegraphics[width=\linewidth]{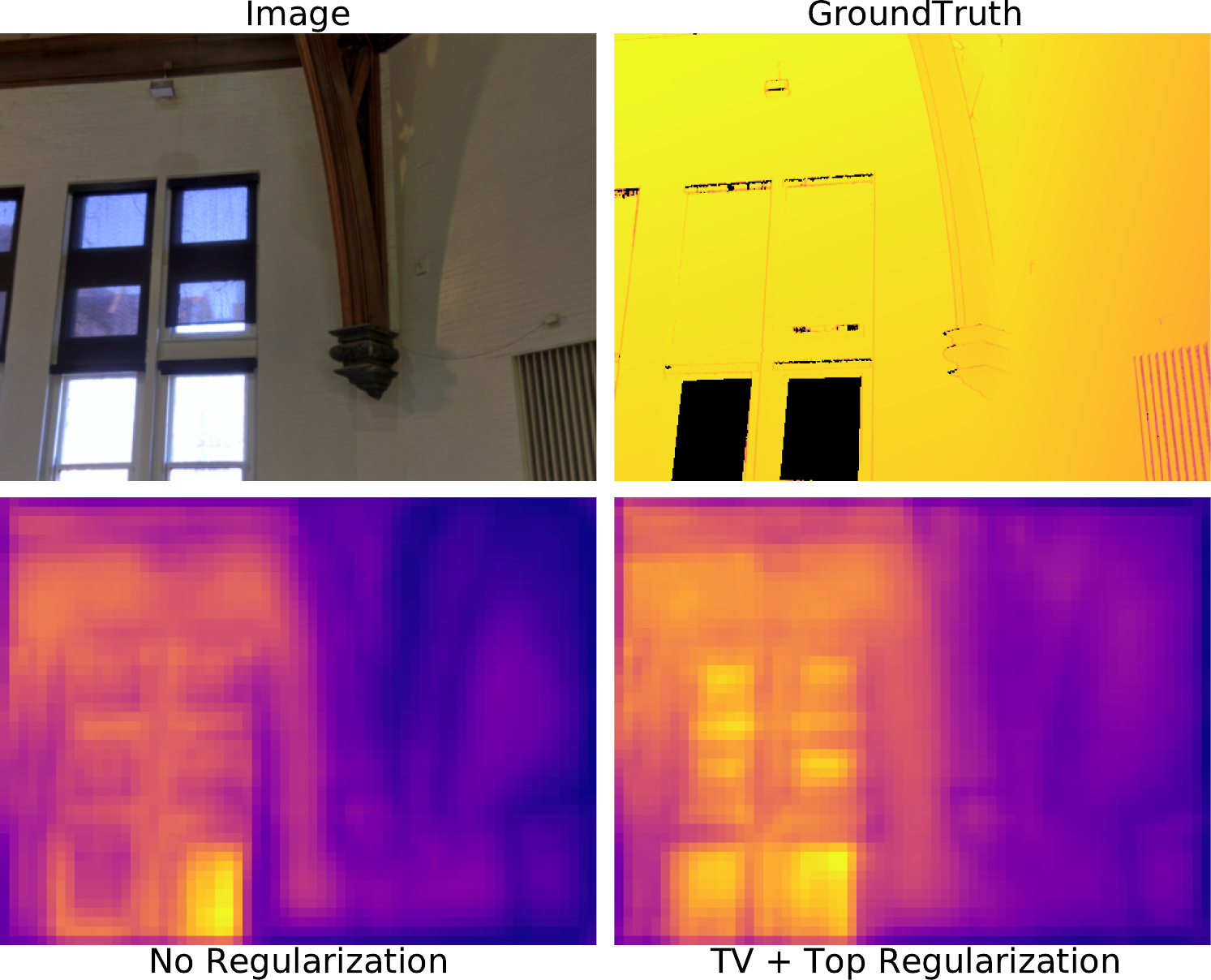}
    \end{subfigure}
    \caption{\textbf{Regularization Effects on DenseDepth.} The second row shows internal activations. Our regularized version also keeps concentration on regions and makes foreground and background objects more separable. }
    \label{fig:comparison_densedepth}
\end{figure*}


%% file: 01Intro.tex
\vspace{-1ex} \section{Introduction}
\label{sec:intro}

Dense prediction problems are a class of problems which attempt to infer some value at every pixel in an image.  Examples include semantic segmentation and monocular depth estimation which we consider here, as well as instance segmentation, hierarchical boundary detection, and figure-ground inference.
Convolutional neural network models have proved to be highly effective at solving such tasks, and many common architectures for dense prediction are built using an encoder and decoder stitched together in some variation of a U-Net \cite{Ronneberger2015UNetCN}.  Despite their success, neural networks have drawbacks such as expensive training (both in terms of time and power consumption) and typically requiring large amounts of data \cite{schwartz_green_2020}.  Regularization using some form of prior knowledge about a problem can be beneficial in reducing the amount of data or the number of iterations needed to train a network, and can also improve the generalization of the network \cite{Goodfellow-et-al-2016}.

Dense prediction tasks on natural images typically ask a network to infer properties of regions of an image corresponding to individual objects.  This partitions an image into relatively few components where pixels in a shared component are treated in a similar manner, at least in the output of the network.  Viewing the output of the task as a function on pixels, this means we expect a function (not necessarily smooth) with few local extrema corresponding to the regions of the image that are most or least prominent in the prediction task.  Total variation regularization \cite{Cands2006RobustUP} is commonly used in imaging tasks where the output is expected to be piece-wise constant.  However, another natural point of view which is agnostic to the smoothness of the function is to penalize the number of maxima or minima directly using level set topology.

We introduce a regularization method based on the super-level set topology of a function which encourages few local maxima in the output of a dense prediction problem.  This method is based on persistent homology \cite{edelsbrunner2000topological} and joins a growing body of topological regularization methods finding use in machine learning \cite{leygonieFrameworkDifferentialCalculus2019,topologyLayerMachine2020,carrierePersLayNeuralNetwork2020,kimEfficientTopologicalLayer2020,hofer_learning_2019}. We find that this super-level set topology is not only important in the output of the network, but also in the internal activations of the decoder portion of the network.  We demonstrate that topological regularization of the internal activations of the network during training leads to faster convergence and improved inference results on both a semantic segmentation task and a monocular depth estimation task.






%% file: 02related_work.tex
\vspace{-1ex} \section{Related Work}
\label{sec:related}

\noindent \textbf{Semantic Segmentation} is a common task in image processing and it requires a dense prediction on each pixel of its corresponding classification. Convolutional neural networks (CNNs) \cite{LeCun1998GradientbasedLA} have shown great capability in solving semantic segmentation problems, and many architectures and methods have been introduced. Region based methods such as R-CNN \cite{Girshick2014RichFH}, and its derivative, Mask R-CNN \cite{He2020MaskR} use a CNN as a feature extractor for region proposals and refine from bounding boxes to semantic segmentation masks. FCN \cite{Shelhamer2017FullyCN} demonstrates pixel-to-pixel prediction capabilities, U-Net  \cite{Ronneberger2015UNetCN} shows great performance on medical images and DeepLab \cite{Chen2018DeepLabSI} proposes atrous spatial pyramid pooling (ASPP) to handle objects of multiple scales. In recent works people are also interested in finding ways of learning semantic representations in an unsupervised manner. Zhang \etal \cite{Zhang2020SelfSupervisedVR} show a way of bootstrapping semantic representation learning via primitive hierarchical grouping such as edges. Aside from development in model architectures and learning schemes, another important line of work seeks to regularize networks to produce high-quality and robust semantic results. Jia \etal \cite{Jia2019ARC} shows that integrating the total variation regularization  \cite{Cands2006RobustUP} to deep convolutional neural networks can both improve the quality of segmentation and its robustness to noises. The key difference of this work to Jia \etal's is that they only regularize the last softmax layer while this work uses topology to regularize internal activations, not just the last layer.

\noindent \textbf{Monocular Depth Estimation} is another topic of interest in the area of dense predictions. There is a wide range of contributions to learning depth from stereo images. The goal of monocular depth estimation, however, is to predict depth from a single RGB image. Eigen \etal \cite{Eigen2014DepthMP} initiated the idea of using deep neural networks to estimate depth without using superpixelation. More recent work follow the step of improving the capability of convolutional neural networks. Alhashim \etal  \cite{Alhashim2018HighQM} propose the DenseDepth model that leverage transfer learning, to use ImageNet \cite{Russakovsky2015ImageNetLS} pretrained DenseNet \cite{Huang2017DenselyCC} as encoders to extract features and a decoder composed of basic convolution layers to reconstruct depth. Lee \etal \cite{Lee2019FromBT} proposed the BTS model that use local planar guidence layers to further improve the encoder-decoder scheme. \\
\indent Aside from the supervised manner of training on one specific dataset, Lasinger \etal \cite{Lasinger2020TowardsRM} demonstrate a zero-shot technique to mix multiple data sets, even with incompatible annotations, which can improve the generalization skill of the networks. Outside of the family of convolutional neural networks, Ranftl \etal \cite{Ranftl2021VisionTF} extends the vision transformers \cite{Dosovitskiy2021AnII} to monocular depth estimation, and achieves better results than CNNs. This work will focus on CNNs, but discussions on vision transformers will be a promising direction for future work. Depth maps are intrinsically equipped with one nice property: they only have a small number of local extrama corresponding to object instances in the images. This work will leverage this property and show that a topological regularization on internal activations can also improve monocular depth estimation tasks.\\
\textbf{Topology and Neural Networks.}
The key contribution of this work is to introduce a method of topological regularization for dense prediction problems which is built on persistent homology.  Persistent homology \cite{edelsbrunner2000topological} is a an algebraic signature of filtrations of topological originating in topological data analysis which measures the robustness of topological features such as connected components and holes in a topological space.  There has been a recent broad effort to develop persistent homology-based losses and regularization terms in a variety of contexts, including the development of application-specific losses \cite{rbgsurface,chenTopologicalRegularizerClassifiers2018,hofer_graph_2020,shit_cldice_2021} as well as general implementations and theory to make this tool increasingly available 
\cite{leygonieFrameworkDifferentialCalculus2019,topologyLayerMachine2020,carrierePersLayNeuralNetwork2020,kimEfficientTopologicalLayer2020,hofer_learning_2019}.  Several applications to computer vision have previously appeared, including a previous application to semantic segmentation by Hu \etal \cite{hu_topology-preserving_2019}.  In contrast to Hu \etal our method does not use any form of ground truth in the regularization, or an expensive Wasserstein distance computation.  Our regularization scheme is similar to those suggested for the use in training generative adversarial networks on images of objects by Br\"uel-Gabrielsson \etal \cite{topologyLayerMachine2020}, and we leverage the implementation provided in their work.\\ \indent 
A particularly novel aspect of our work is the application of topological regularization to the internal activations of a network, and, to the best of our knowledge, this is the first work to demonstrate the utility of doing so.  Typically, topological losses or regularization terms are applied to the output of the network, but as we see, dense prediction tasks manifest similar topology in earlier activations as well.  Our approach is inspired by recent work by Naitzat \etal \cite{naitzat2020} that has shown that trained neural networks tend to simplify topology in internal activations, although their topological construction uses Rips complexes built on entire data sets as they pass through a network, and our work focuses on level set filtrations of single inputs.

%% file: 03tda.tex
\vspace{-1ex} \section{Topological Regularization}
\label{sec:tda}
\begin{figure}[b]
    \centering
    \includegraphics[width=0.24\linewidth]{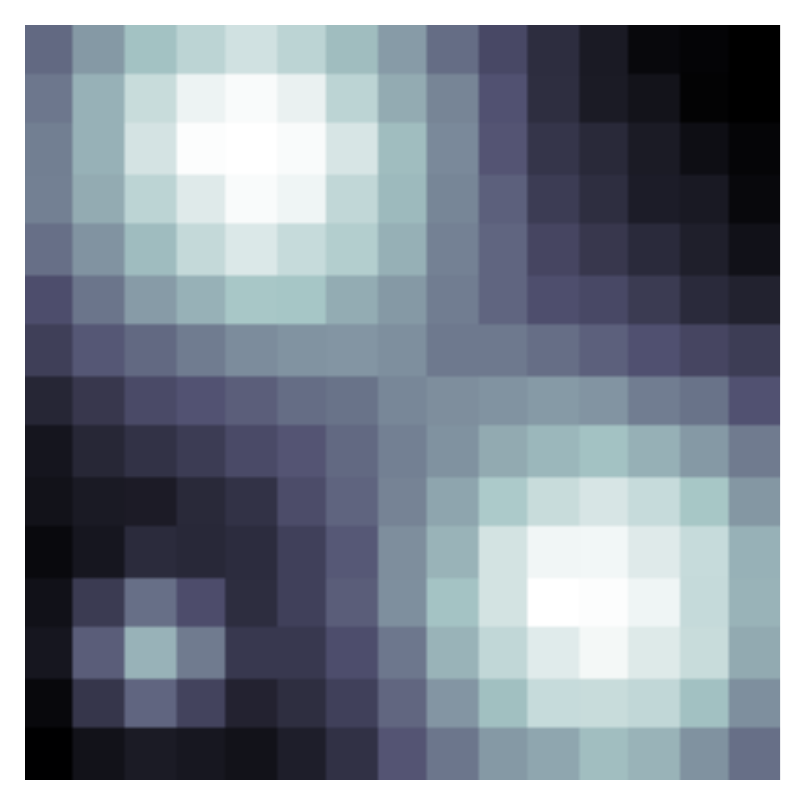}
    \frame{\includegraphics[width=0.24\linewidth]{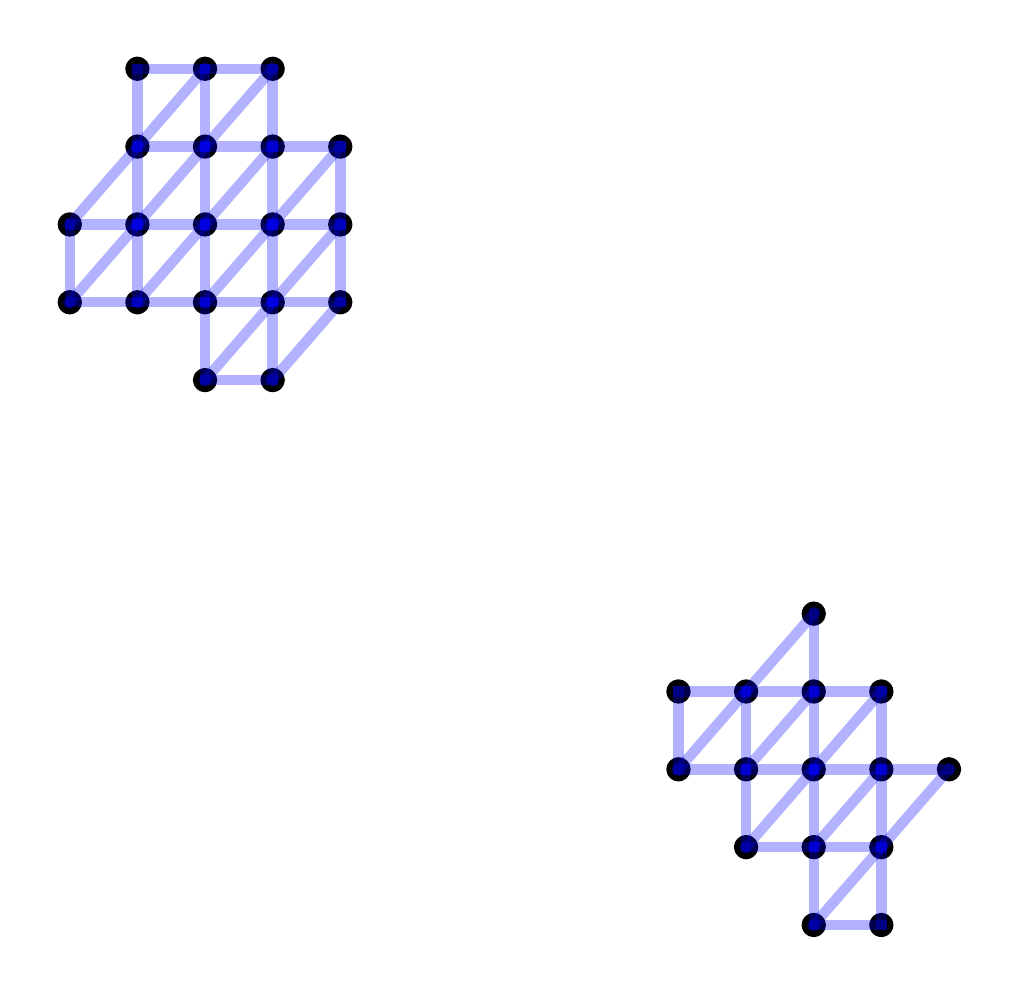}}
    \frame{\includegraphics[width=0.24\linewidth]{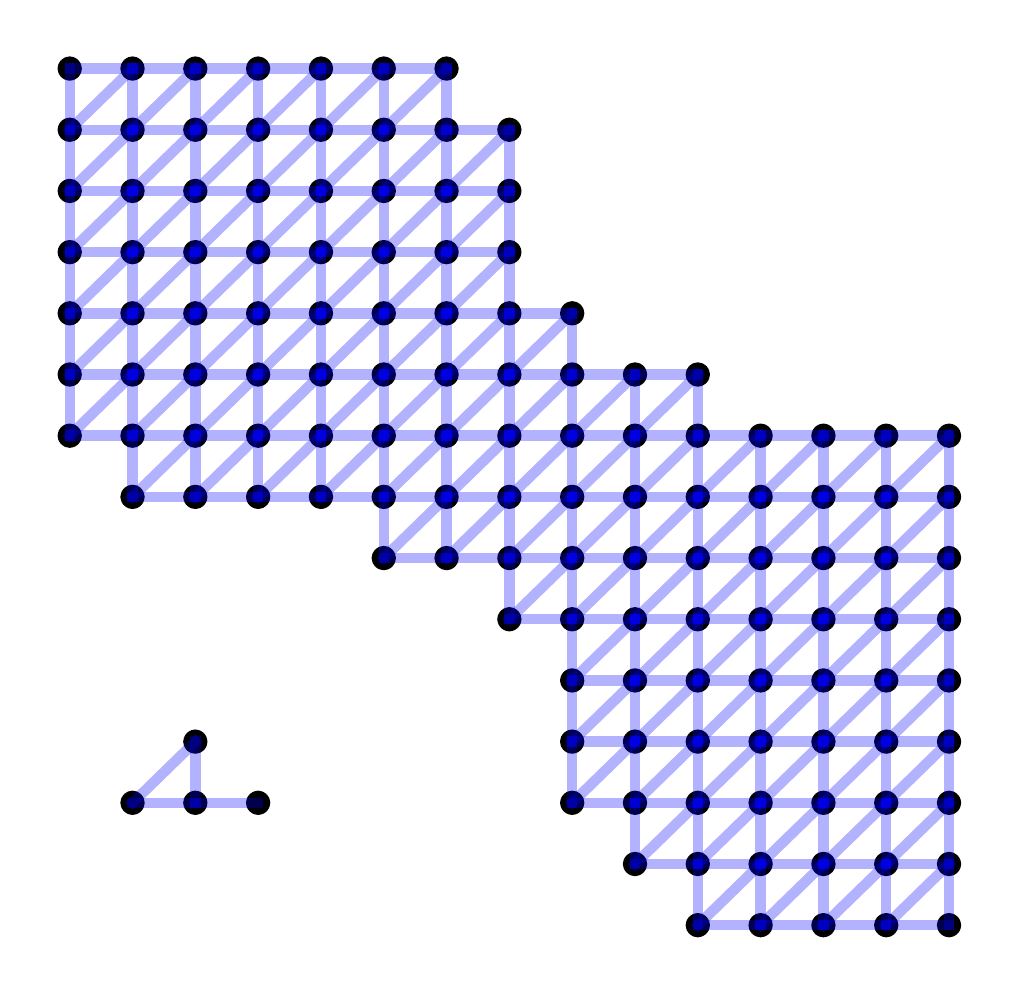}}
    \frame{\includegraphics[width=0.24\linewidth]{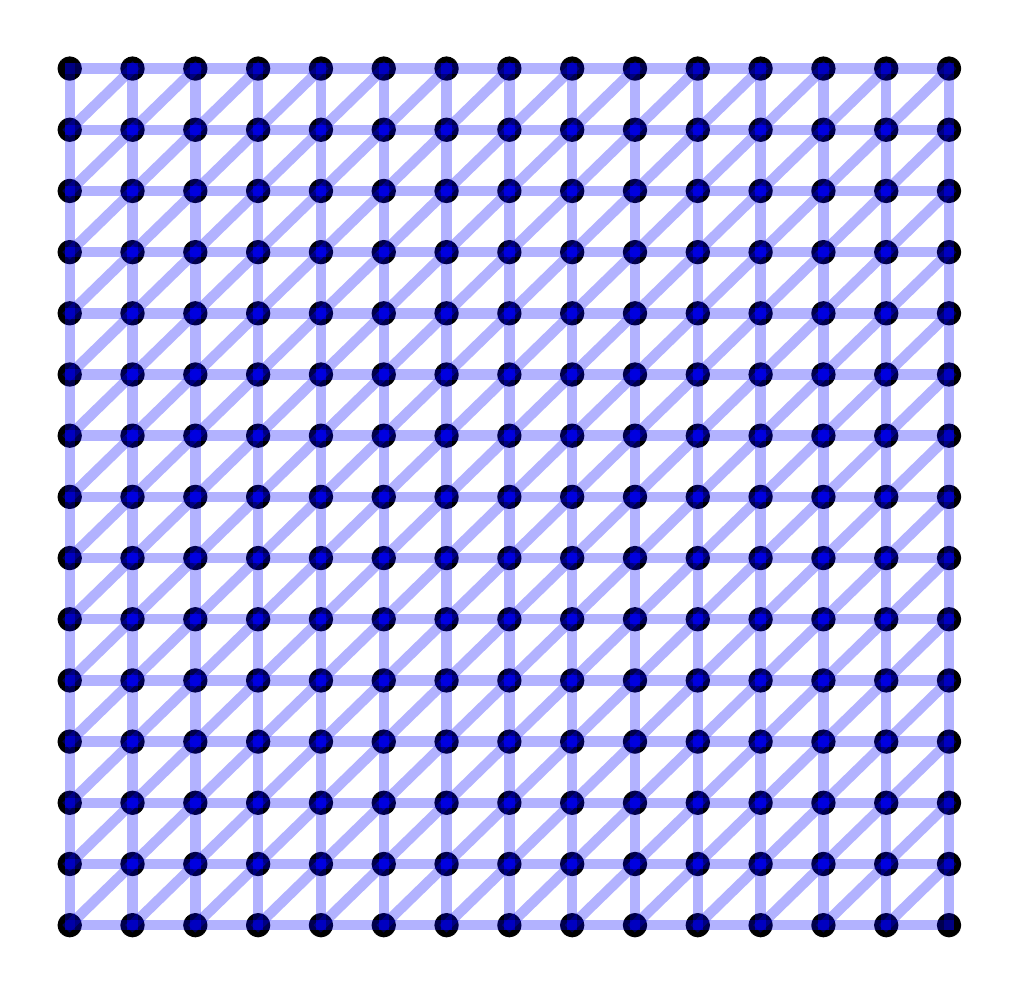}}
    \caption{Left to right: function $f$ of pixel values over an example image where light pixels have higher value than dark. Then a sequence of simplicial complexes obtained from super-level sets: $f^{-1}([0.85,\infty))$, $f^{-1}[0.45,\infty))$, and $f^{-1}([0,\infty))$.}
    \label{fig:levelset_ex}
\end{figure}

In this section we provide a brief introduction to the persistent homology of super-level set filtrations, which is the key construction in our topological regularization scheme.  For surveys of of persistent homology which provide broader context and additional detail see \cite{carlssonTopologyData2009,carlsson_topological_2014}, and for additional background on computing persistent homology see \cite{otterRoadmapComputationPersistent2017}. The topological spaces we use are all thought of as subsets of a rectangle, discretized using the Freudenthal triangulation so that the vertex set of the triangulation is arranged to correspond to the pixel grid of an image.  This gives a combinatorial representation of the rectangle as a \emph{simplicial complex} $\mathcal X = \mathcal X_0 \cup \mathcal X_1 \cup \mathcal X_2$, consisting of a vertex set (0-simplices) $\mathcal X_0$, an edge set (1-simplices) $\mathcal X_1 \subset \mathcal X_0 \times \mathcal X_0$, and a set of triangles (2-simplices) $\mathcal X_2 \subset \mathcal X_0 \times \mathcal X_0 \times \mathcal X_0$.  We will denote $k$-simplices as $(x_0,\dots,x_k), x_i\in \mathcal X_0$.

Images can either be input or output images of a neural network (with one or more channels), or the internal activations of a neural network on a particular image (typically many channels). A super-level set filtration uses a real-valued function $f$ which takes values over each pixel, in our case we take the 2-norm of the channel values over each pixel.  This function can be extended from a function on pixel values (the vertex set of $\mathcal X$) to higher-dimensional simplices using a lower-star filtration $f(x_0,\dots,x_k) = \min_{i=0,\dots,k} f(x_i)$.
A filtration is a nested sequence of topological spaces $\{\mathcal X_a\}$.  Super-level set filtration for a function $f$ extended to a whole simplicial complex use $\mathcal X_a = f^{-1}([a,\infty))$, which satisfies the inclusion condition $\mathcal X_a \subseteq \mathcal X_b$ if $a > b$.  An example image with several snapshots of the associated super-level set filtration can be seen in \cref{fig:levelset_ex}.

\begin{figure}[b]
    \centering
    \includegraphics[width=\linewidth]{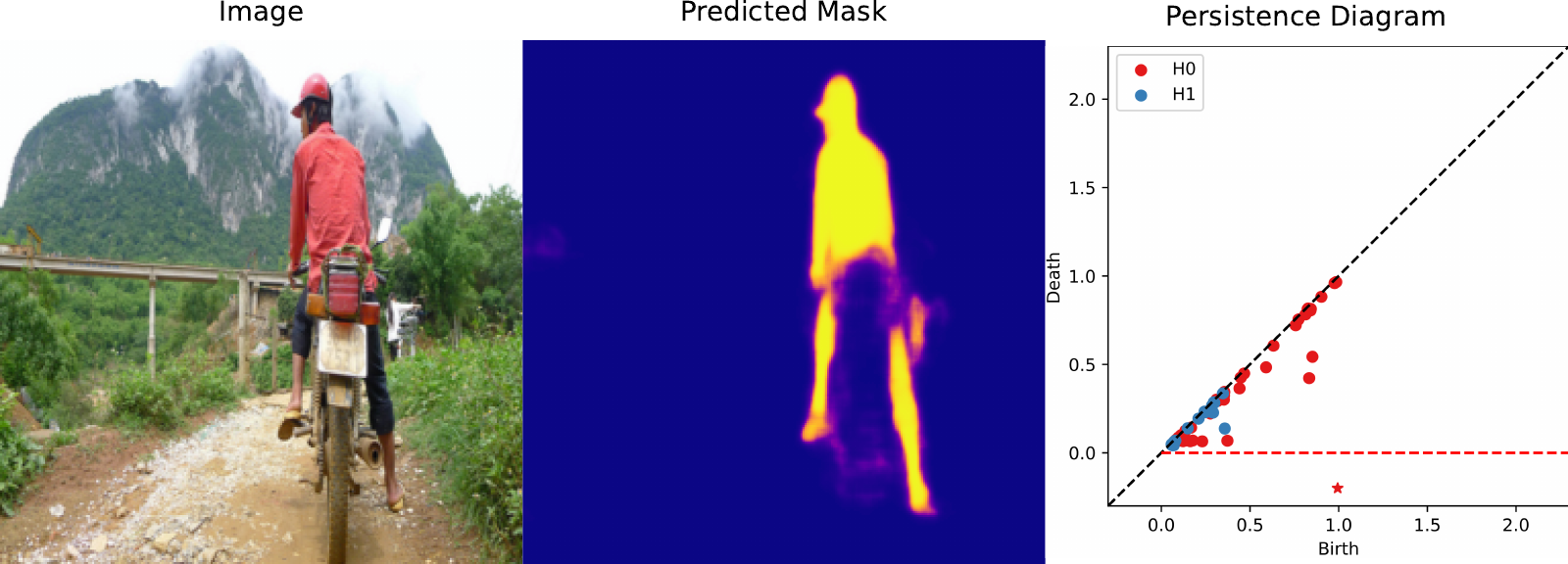}
    \caption{Left to right: an input image from the COCO data set \cite{Lin2014MicrosoftCC}; semantic segmentation mask; persistence diagram of the mask. Each point in the persistence diagram is a persistence pair: red points are homology in dimension 0, and blue points are homology in dimension 1.  There is a single red point below the dashed red line representing the single connected component present at the end of the filtration.}
    \label{fig:levelset_pd}
\end{figure}

Persistent homology is an algebraic invariant of filtrations which summarizes how topological features such as connected components and holes appear and disappear as the filtration parameter increases.  Homology is computed first starting with chain complexes $\{C_k(\mathcal X)\}_{k\ge 0}$ which are vector spaces with basis vectors for each $k$-simplex in $\mathcal X$, and connected by boundary maps $\partial_k:C_k\to C_{k-1}$ defined (for simplicial complexes) as $\partial_k (x_0,\dots,x_k)\mapsto \sum_i (-1)^k(x_0,\dots,\hat{x}_i,\dots,x_k)$, where $\hat{x_i}$ denotes the removal of the vertex $x_i$ to obtain a $k-1$ simplex from the $k$-simplex, and $\partial_0 = 0$.  These boundary maps are differentials: $\partial_k \partial_{k+1} = 0$ for all $k\ge 0$.  

Homology in dimension $k$ is the quotient vector space $H_k = \ker \partial_k / \img \partial_{k+1}$. The rank of homology in dimension $0$ counts the number of connected components of $\mathcal X$, in dimension $1$ counts the number of holes, and, generally, in dimension $k$ counts $k$-dimensional voids.  Homology is a functor, meaning that the inclusions $\mathcal X_a \subseteq \mathcal X_b$ have associated linear maps $F_{k;a,b}:H_k(\mathcal X_a) \to H_k(\mathcal X_b)$ which are useful in determining how features in $\mathcal X_a$ map to features in $\mathcal X_b$.  \emph{Persistent homology} $PH_k(\{\mathcal X_a\})$ describes how vectors first appear in the co-kernel of a map and then survive the application of maps induced by inclusion until eventually entering the kernel.  Persistent homology is characterized up to isomorphism \cite{ZCComputingPH2005,ZZtheory2010} by birth-death pairs $PH_k(\{\mathcal X_a\}) = \{(b_i,d_i)\}$ where the pair $(b_i,d_i)$ describes the birth of a new vector at parameter $b_i$ and death of its image at parameter $d_i$, often visualized plotted in the plane as a \emph{persistence diagram} -- see \cref{fig:levelset_pd} for an example.  Pairs with well-separated birth and death are robust to perturbation of function values \cite{cohen-steinerStabilityPersistenceDiagrams2007}, and points with nearby births and deaths are typically considered topological noise.

Each birth or death in persistent homology has a subgradient, one element of which is obtained by associating the birth or death to a particular simplex that changed the rank of homology at the parameter the birth or death occurred.  While this mapping is not generally unique, algorithms for computing persistent homology produce a choice of one-to-one mapping - see Br\"uel-Gabrielsson \etal \cite{topologyLayerMachine2020} for additional details.  The particular regularization we use is in the family of functionals based on algebraic functions of the birth-death pairs \cite{adcock_ring_2016}.  Specifically, we penalize all but the $k$ longest birth-death pairs in dimension 0:
\begin{equation}
    \mathcal L_\text{Topology}(\{b_i,d_i\}) = \sum_{i > k} |d_i - b_i|^2.
    \label{eqn:top}
\end{equation}
This encourages at most $k$ local maxima in the function $f$ over the image channels.

\noindent \textbf{Computation. } To apply our topological loss we use the TopologyLayer PyTorch package \cite{topologyLayerMachine2020} modified to use the union-find algorithm \cite{tarjan_class_1979} to compute $PH_0$, which runs in $O(m\alpha(m))$ time, where $m$ is the number of edges in the simplicial complex and $\alpha$ is the inverse Ackermann function.  To compute persistent homology in higher dimensions it is necessary to perform a factorization of boundary matrices which can be achieved in matrix multiplication time \cite{ZZmatmultime2011}.  Standard implementations are asymptotically cubic in the number of simplices, but sparsity in boundary matrices often make this bound pessimistic \cite{otterRoadmapComputationPersistent2017}.  An advantage of our method is that the size of the spaces we use to regularize internal activations of a network are an order of magnitude smaller than the size of the output space, significantly reducing the cost of using a topological penalty.

%% file: 04internal_activations.tex
\section{Topology of Internal Activations}

\begin{figure*}
    \centering
    \includegraphics[width=0.925\linewidth]{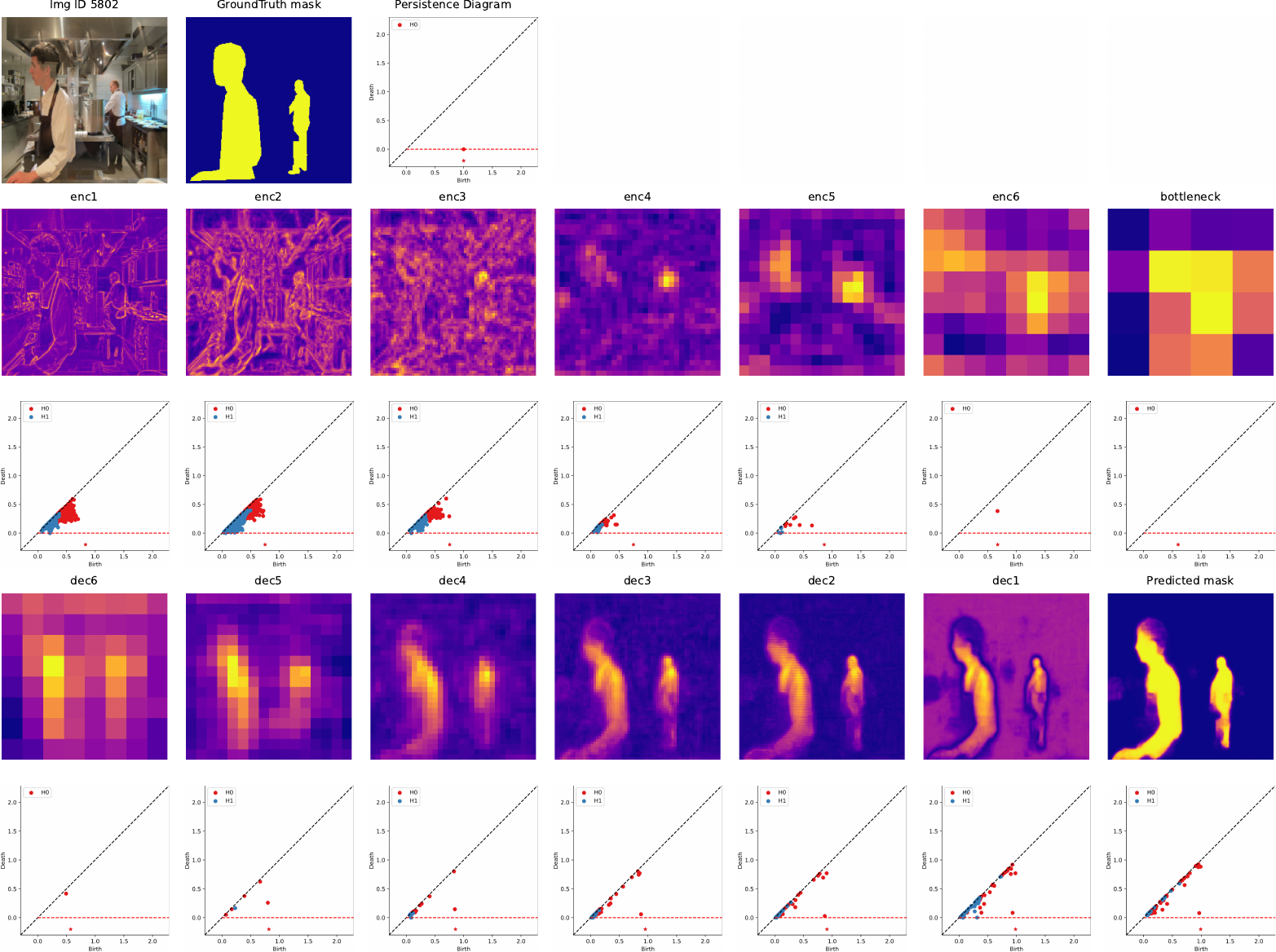}
    \caption{\textbf{Visualizations and Persistence Diagrams of Internal Activations.} The first row shows the input image, the ground truth segmentation mask, and the mask's persistence diagram. The 2nd and 4th row show the visualizations of the magnitude of internal activations. On the 3rd and 5th row, each figure shows the persistence diagrams of the corresponding internal activations above.  Points further from the diagonal are more robust features. Both visualizations and diagrams show that the level-set topology can emerge as early as in the 4th encoder layer, and it remains nearly consistent in the decoder layers.  No topological regularization is used on this example.}
    \vspace{-2ex}
    \label{fig:internal}
\end{figure*}

In this section, we provide experimental results which demonstrate that topology can appear in the internal activations of trained neural networks. We train a convolutional neural network on binary semantic segmentation task which assigns values to each pixel indicating whether this pixel is part of a human torso or not. 

\noindent \textbf{Model architecture.} We use the U-Net \cite{Ronneberger2015UNetCN} architecture, which is a popular baseline for segmentation tasks. Our U-Net adopts 6 phases of encoders and 6 phases of decoders, where each block consists of two layers of convolutions, batch normalizations, and ReLU activations. \Cref{fig:internal} shows the flow of the network, along with visualizations of each layer, and the network progresses as \verb+Input+ $\rightarrow$ \verb+enc1+ $\rightarrow \cdots \rightarrow$ \verb+enc6+ $\rightarrow$  \verb+bottleneck+ $\rightarrow$ \verb+dec6+ $\rightarrow \cdots \rightarrow$ \verb+dec1+ $\rightarrow$ \verb+Output+ with skip connections. 

\noindent \textbf{Dataset.} We use the COCO dataset \cite{Lin2014MicrosoftCC}, specifically COCO-2014, which contains 83K training images and 41K validation images. From its provided semantic segmentation annotations, we make a subset where all images contain human annotations and further process them to only annotate pixels as human/non-human. This commonly used data set contains human objects and \cite{Zhao2021UnderstandingAE,Hu2018ExploringSA} have recently studied bias in captions.  However, our semantic segmentation task does not use these potentially problematic labels.

\noindent \textbf{Training objectives and hyper-parameters.} The training objective is the MSE loss
\begin{equation}
    \mathcal L_\text{MSE}(y, \hat{y}) = \sum_{i=1}^h \sum_{j=1}^w (y_{i,j} - \hat{y}_{i,j})^2
\end{equation}
where $h$ and $w$ are the height and width of the image. We train the network for 100 epochs with initial learning rate of 0.01. We use SGD as the optimizer with learning rate decaying by half every 40 epochs.

\noindent \textbf{Experimental results.} As discussed in \Cref{sec:tda}, since each intermediate layer $h \in \mathbb R^{h \times w \times c} $ has more than one channels, e.g. $c > 1$, there's a natural and differentiable projection $\pi$ that send the $c$-dimensional vector at each pixel to a real value, in order to apply super-level set filtration and compute persistence diagrams. The projection $\pi: R^{h \times w \times c} \rightarrow R^{h \times w \times 1}$ is defined as, for each pixel $(i,j)$, 
\begin{equation}
    \pi(\mathbf{h}_{i,j}) = ||\mathbf{h}_{i,j}||_2
    \label{eqn:projection}
\end{equation}

We evaluate on a completely trained network by visualizing the projection, by \Cref{eqn:projection}, of each internal activation and computing persistence diagrams. As shown in the example of \Cref{fig:internal}, the desired ground-truth mask is consisted of two connected components since there are two human torsos in the input image. 

As we follow the steps of this trained convolutional neural network by looking at its internal activations, we can see that the first 3 encoder layers, \verb+enc1+, \verb+enc2+, and \verb+enc3+, present low-level edge structures, and their corresponding diagrams do not reveal any simple topology. Starting from the 4th encoder layer both the visualizations and diagrams demonstrate that topology starts to simplify and two connected components emerge seen as two $PH_0$ pairs well separated from the diagonal.  

Once we proceed to the decoder layers, both visualizations and persistence diagrams illustrate that the topology remains fairly consistent through all decoder layers and the output layer.  In \Cref{fig:internal} these two components eventually form the segmentation masks for the two people in the image.  In other images the number of robust components in the internal activations typically agrees with the number of connected components in the final segmentation mask. More examples can be viewed in the supplemental materials.  This provides a natural suggestion of explicitly regularizing the internal activations throughout the training stage.

\begin{figure}[h]
    \centering
    \includegraphics[width=0.9\linewidth]{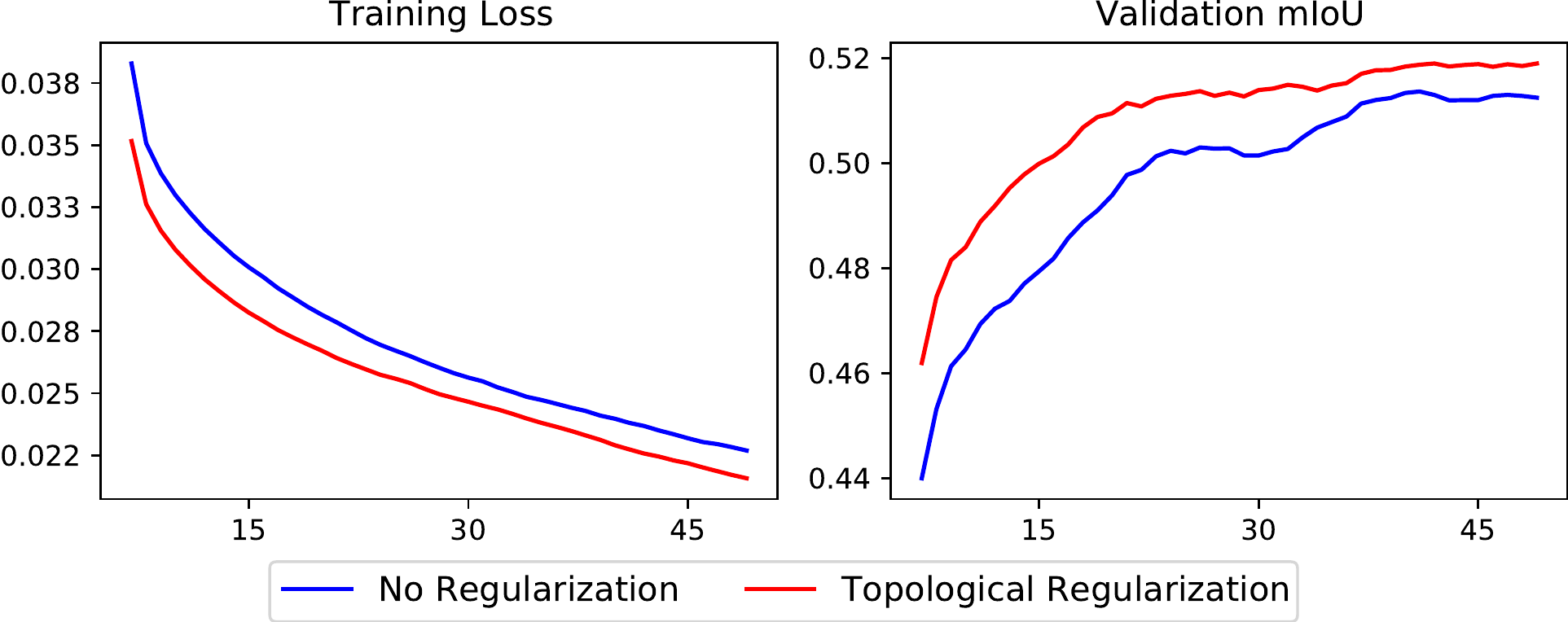}
    \caption{Convergence Improvement by Topology Regularizer}
    \vspace{-4ex}
    \label{fig:semantic_segmentation_convergence}
\end{figure}

\noindent \textbf{Topological Regularization on binary Semantic Segmentation.} We experimented using the same architecture as described above and trained the networks for 50 epochs with cosine learning rate schedule, but varying the choice of regularization. We regularize \verb+dec4+ with $k = 8$ to penalize more than 8 connected components on this intermediate layer. As shown in \Cref{tab:semantic_segmentation_performance}, there is a improvement on \verb+mIoU+, e.g., mean Intersection-over-Union accuracy, when we regularize the second decoder layer with the proposed topological regularizer. As shown in \Cref{fig:semantic_segmentation_convergence}, there are also improvements on convergence speed with the assistance of the proposed regularization.

The next section will show how this explicitly regularization can improve the convergence and test benchmarks on several architectures on depth prediction tasks. 

\begin{table}[h]
    \centering
    \vspace{1ex}
    \small
    \begin{tabular}{c|c}
    \toprule
         & mIoU (\%)  \tstrut \bstrut \\
         \hline
         No Regularization & 51.64 \tstrut \bstrut \\
         Topological Regularization & \textbf{52.54} \tstrut \bstrut\\
    \bottomrule
    \end{tabular}
    \caption{Performance Improvement by Topology Regularizer}
    \vspace{-4ex}
    \label{tab:semantic_segmentation_performance}
\end{table}

%% file: 05experiment.tex
\section{Experiments}
\label{sec:experiments}
In this section, we explore the performance improvement on monocular depth estimation with different level of regularizations. We start with the base U-Net architecture, and compare the performance without regularization, with only total variation regularization, and with both total variation and topological regularization. We'll show that both regularization on internal activations assist the convergence and performance of the U-Net model. Subsequently, we experiment with a recent state-of-the-art architecture DenseDepth to demonstrate the versatility of our proposed regularization. 

\subsection{Training Objectives}
Our proposed training objective is a weighted sum of three loss functions and two regularization terms:
\begin{equation}
\small
\begin{aligned}
    \mathcal L(y, \hat{y}) &= \lambda_\text{d} \cdot \mathcal L_\text{depth}(y, \hat{y}) + \lambda_\text{g} \cdot \mathcal L_\text{gradient} (y, \hat{y}) + \lambda_\text{s} \cdot \mathcal L_\text{SSIM} (y, \hat{y}) \\
    & \quad + \lambda_\text{tv} \mathcal \cdot \mathcal  L_\text{TotalVariation}(\hat{h}^{(a)}) + \lambda_\text{top} \cdot  \mathcal L_\text{Topology}(\hat{h}^{(b)})
\end{aligned}
    \label{eqn:objective}
\end{equation}
where $y$ represents the ground-truth label of depth, $\hat{y}$ represents the predicted depth, $\hat{h}^{(a)}$ and $\hat{h}^{(b)}$ represents some intermediate layers of the network. Each loss term is defined as follows,

\noindent \textbf{Depth Loss. } We use the RMSE loss in log scale which empirically converges faster than L1 or L2 loss.
\begin{equation}
\small
    \mathcal L_\text{depth}(y, \hat{y}) = \sqrt{\sum_{i=1}^h \sum_{j=1}^w (\log y_{i,j} - \log \hat{y}_{i,j})^2}
\end{equation}

\noindent \textbf{Gradient Loss. } The horizontal and vertical image gradients, $\nabla_\parallel$ and $\nabla_\perp$, are computed by a Sobel filter \cite{Sobel1990AnI3,kanopoulos1988design}. This further helps align the edges of ground-truths and predictions. 
\begin{equation}
\small
\begin{aligned}
     \mathcal L_\text{gradient}(y, \hat{y}) = \sum_{i=1}^h \sum_{j=1}^w  \big(\left|\nabla_\perp y_{i,j} - \nabla_\perp \hat{y}_{i,j}\right| + \left|\nabla_\parallel y_{i,j} - \nabla_\parallel  \hat{y}_{i,j}\right|\big)
\end{aligned}
\end{equation}

\noindent \textbf{Structural Similarity Loss. } This loss uses the Structure Similarity Index Measure (SSIM) \cite{Wang2004ImageQA} which is shown to be a good loss term for depth estimation tasks \cite{Godard2017UnsupervisedMD}. 
\begin{equation}
\small
    \mathcal L_\text{SSIM}(y, \hat{y}) = \frac{1 - \SSIM(y, \hat{y})}{2}
\end{equation}

\noindent \textbf{Total Variation Regularization. } It's often used in imaging tasks where the expected output is piece-wise constant. We apply this regularization term to the last layer $\hat{h}^{(a)}$ before the output layer. 
\begin{equation}
\small
\begin{aligned}
    \mathcal L_\text{TotalVariation}(\hat{h}^{(a)}) = & \sum_{k=1}^c \sum_{i=1}^{h-1} \sum_{j=1}^w \left(\hat{h}^{(a)}_{i+1, j, k} - \hat{h}^{(a)}_{i, j, k}\right)^2 \\ 
    &+ \sum_{k=1}^c \sum_{i=1}^{h} \sum_{j=1}^{w-1} \left(\hat{h}^{(a)}_{i, j+1, k} - \hat{h}^{(a)}_{i, j, k}\right)^2
\end{aligned}
\end{equation}

\noindent \textbf{Topological Regularization. } We apply this regularizer to the second decoder layer  $\hat{h}^{(b)}$. We first use the projection described by \Cref{eqn:projection} to project this internal activation to $\tilde{h} \in \mathbb R^{h \times w}$ and compute its birth-death pairs $(b_i, d_i)$ using super-level set filtration and persistent homology described in \Cref{sec:tda}. Afterwards, we formulate the loss, given these birth-death pairs, through \Cref{eqn:top}, e.g., $\displaystyle \mathcal L_\text{Topology}(\{b_i,d_i\}) = \sum_{i > k} (d_i - b_i)^2$. For our application, we choose $k = 8$ to penalize internal activations to have more than 8 connected components or local extrema. The hyperparameter $k$ can be chosen differently, but we choose $k$ to obtain the best performance. 

To achive the best performance, we set the weights of these objectives as $\lambda_d = 0.1, \lambda_g = 1.0$ and $\lambda_s = 1.0$. As un-regularized DenseDepth to have clearer topology on internal activations than U-Net (see \Cref{fig:qualitative_comparison}), we choose different weights of $\lambda_\text{tv}$ and $\lambda_\text{top}$ for each. For U-Net, we set $\lambda_\text{tv} = 1.0$ and $\lambda_\text{top} = 0.001$; and for DenseDepth, we set we set $\lambda_\text{tv} = 0.1$ and $\lambda_\text{top} = 0.0001$.

\subsection{Models and Training Details}
\noindent \textbf{U-Net. } We use variants of a ResNet-50 backboned U-Net developed by \cite{Mate:2018} and \cite{Yakubovskiy:2019}. This replaces the encoder part of the original U-Net with a ImageNet pretrained ResNet-50 \cite{He2016DeepRL} feature extractor. We trained the model in three settings, without regularization, with total variation regularization, and with total variation plus topological regularization. For all three training procedures, we set the initial learning rate for parameters of the decoder to 0.03 and we keep the learning rate for parameters of the pretrained ResNet encoder to be 1/10 of that of the decoder. Models are trained for 100 epochs with batch size 16, a cosine learning rate schedule, a momentum of 0.9 and a weight decay of 1e-4. For regularization settings, the total variation regularization is enforced onto the last decoder layer and the topological regularization is enforced onto the second decoder layer. 

\noindent \textbf{DenseDepth.} We use DenseDepth with DenseNet-161 encoder. We trained the model in two settings, without regularization, and with total variation regularization plus topological regularization. To be comparable with scores reported in \cite{Vasiljevic2019DIODEAD}, we use Adam optimizer \cite{Kingma2015AdamAM} with initial learning rate 0.0001, $\beta_1 = 0.9$ and $\beta_2 = 0.999$. Models are trained for 20 epochs with batch size 12 and a cosine learning rate schedule. For regularization settings, we keep it the same as in the U-Net experiments, e.g., the total variation regularization is enforced onto the last decoder layer and the topological regularization is enforced onto the second decoder layer. 

Experiments are trained on two nVidia 2080 Ti's with 11GB memory each. 

\subsection{Data Set and Augmentation Policy}
\noindent\textbf{DIODE} is a data set that provides images, depth maps and surface normals for both indoor and outdoor scenes \cite{Vasiljevic2019DIODEAD}. Data provided are captured at a resolution of 1024 $\times$ 768. It contains 8,574 indoor scenes and 16,884 outdoor scenes for training. The maximum depth captured by the sensor is 350 meters and the minimum depth is 0.6 meters. Both our models take half of the orginal resolution as input, i.e., a resolution of 512 $\times$ 384. The U-Net model produces predictions at a resolution of 512 $\times$ 384, and the DenseDepth model produces predictions at a resolution of 256 $\times$ 192 followed by a 2x upsampling through bilinear interpolations. 

\noindent\textbf{Augmentation policy}
Data augmentation is universally used to reduce over-fitting and can result in better generalization skills. Since the monocular depth estimation tasks aim to predict depth from an entire image, geometric transformations may not be appropriate choices since they may introduce additional distortions that are not natural in depth estimation. We adopt similar data augmentation strategies as \cite{Alhashim2018HighQM}. For geometric augmentations, We only performed random horizontal flips with a probability of 0.5. For photo-metric augmentations, we performed random color jittering with a probability of 0.8, random channel swapping with a probability of 0.5, randomly converting images to gray scale with a probability of 0.2, and a random Gaussian blurring with a probability of 0.5. It's unknown if more hand-crated augmentations or automated augmentation policies \cite{Cubuk2020RandaugmentPA,Cubuk2019AutoAugmentLA} can also help with generalization skills of trained networks on monocular depth estimation and is an interesting topic for future research. 

\begin{table*}
        \centering
        \begin{tabular}{c |  c | c c c c c | c c c }
        \toprule
            & \multirow{ 2}*{\makecell{Level of \\ Regularization}}  & \multicolumn{5}{c|}{lower is better} &  \multicolumn{3}{c}{higher is better} \tstrut \bstrut \\
            \cline{3-10}
             &  & mae & rmse & abs rel & mae $\log_{10}$ & rmse $\log_{10}$ & $\delta_1$ & $\delta_2$ & $\delta_3$   \tstrut \bstrut \\
              \cline{1-10}
           \multirow{ 5}*{U-Net} 
           & None & 4.2776 & 6.5386 & 0.4524 & 0.1706 &  0.2181 & 0.4336 & 0.6778 & 0.8128 \tstrut \bstrut\\
             \cline{2-10}
            & Total Variation & 4.0952 & 6.3145 & 0.4311 & \color{blue}{0.1649} & \color{blue}{0.2103} & 0.4455 & \color{blue}{0.6915} & 0.8184 \tstrut \bstrut \\
             \cline{2-10}
            & Topology &  \color{blue}{4.0548} & \color{blue}{6.2168} & \color{red}\textbf{0.4206} & 0.1651 & 0.2121 & 0.4388 & 0.6914 & \color{red}\textbf{0.8295}\tstrut \bstrut \\
             \cline{2-10}
            & TV + Topology & \color{red}\textbf{4.0138} & \color{red}\textbf{6.2044} & \color{blue}{0.4269} & \color{red}\textbf{0.1614} & \color{red}\textbf{0.2069} & \color{red}\textbf{0.4565} & \color{red}\textbf{0.7020} & \color{blue}{0.8232} \tstrut \bstrut \\
        \midrule 
        \midrule
           \multirow{ 5}*{DenseDepth} 
           & None & 3.6554 & 5.9900 &  \color{red}\textbf{0.3648} &  0.1660 &  0.2452 &  0.5088 & 0.7481 & 0.8625 \tstrut \bstrut \\
           \cline{2-10} 
           & Total Variation & \color{blue} 3.5073 & \color{blue} 5.5763 & 0.3922 & \color{blue}0.1427 & \color{blue}0.1884 & \color{blue} 0.5151 & 0.7444 & 0.8665 \tstrut \bstrut \\
           \cline{2-10}
           & Topology &  3.5857 &   5.7030 & 0.3921 & 0.1435 & 0.1887 & 0.5006 & 0.7418 & \color{blue}0.8668 \tstrut \bstrut \\
           \cline{2-10}
           & TV + Topology &   \color{red} \textbf{3.4065} &  \color{red}\textbf{5.4196} & \color{blue} 0.3908 &  \color{red}\textbf{0.1395} &  \color{red}\textbf{0.1849} &  \color{red}\textbf{0.5197} & \color{red} \textbf{0.7582} &  \color{red} \textbf{0.8745} \tstrut \bstrut \\
         \bottomrule
        \end{tabular}
        \caption{\textbf{Quantitative Comparison.} UNet and DenseDepth Performance on DIODE. Each model is compared internally with varying regularization choices. Numbers in {\color{red} Red} indicate the best score whereas in {\color{blue} Blue} shows the second best.}
        \vspace{-2ex}
        \label{tab:performace}
\end{table*}

\subsection{Qualitative comparison}
\begin{figure}
    \centering
    \begin{subfigure}[b]{0.24\textwidth}
    \includegraphics[width=\linewidth]{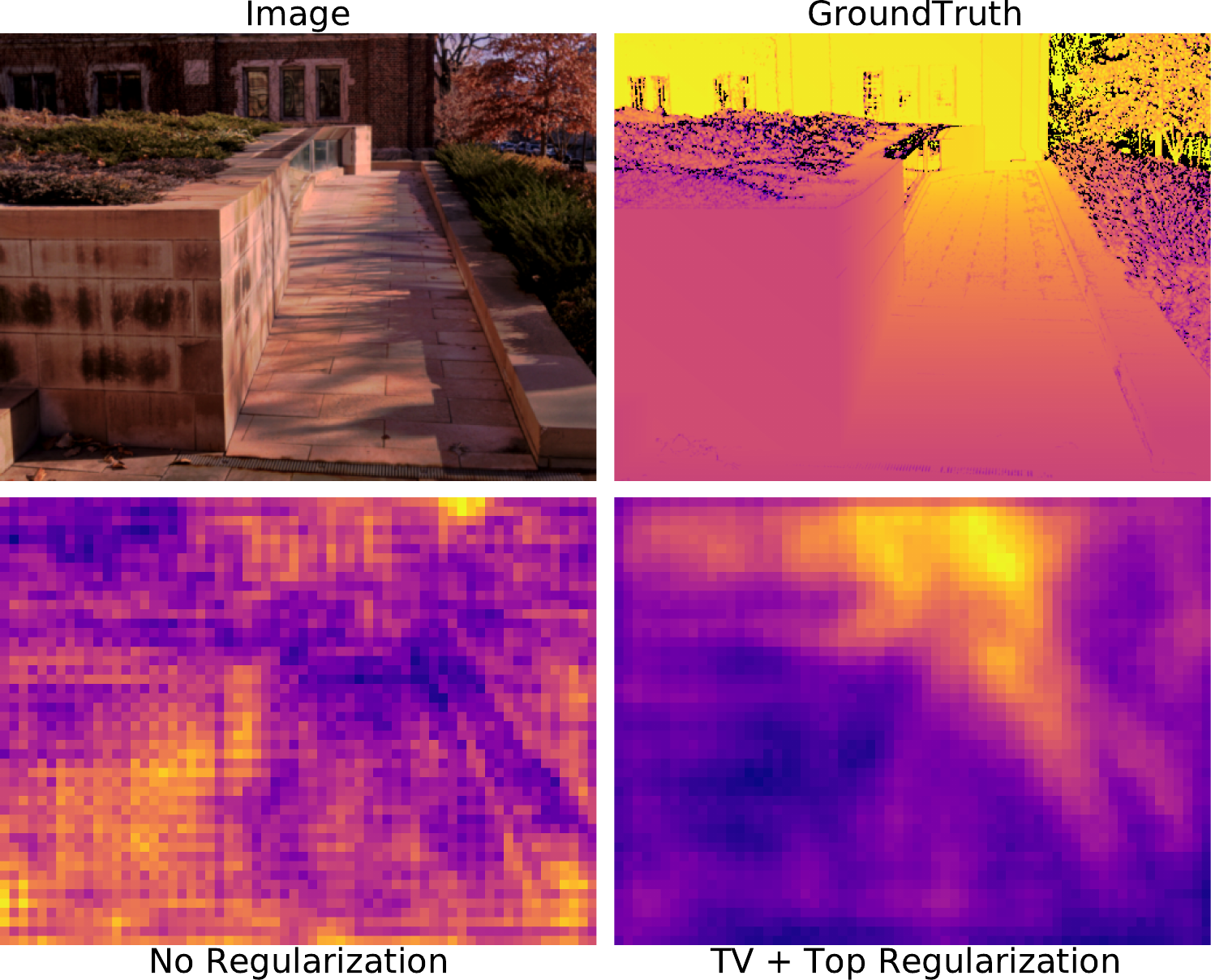}
    \caption{Effects on U-Net.}
    \label{fig:unet_comparison}
    \end{subfigure}
    \begin{subfigure}[b]{0.24\textwidth}
    \includegraphics[width=\linewidth]{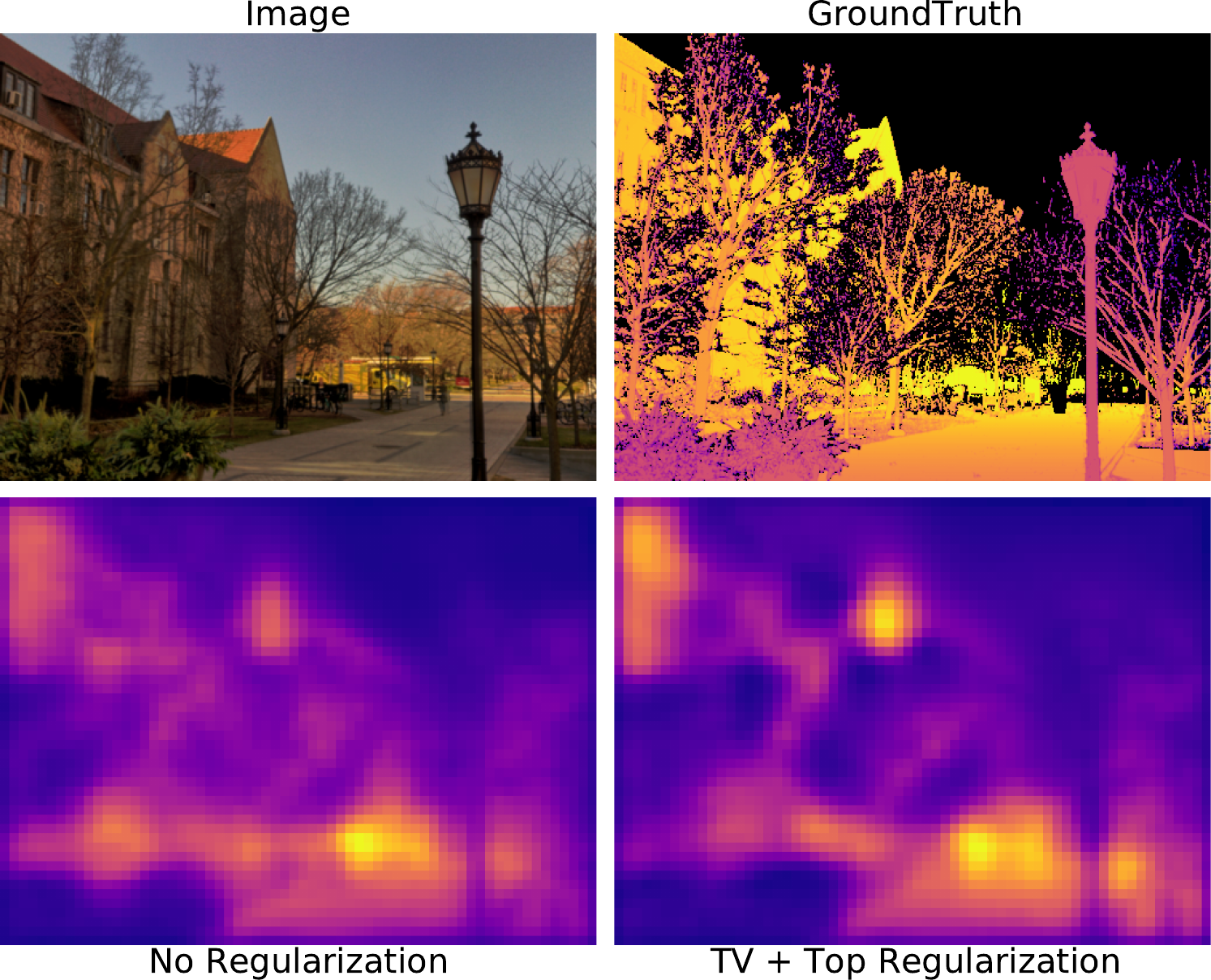}
    \caption{Effects on DenseDepth}
    \label{fig:densedepth_comparison}
    \end{subfigure}
    \caption{\textbf{Effects of Topological Regularization on Interval Activations.} For each architecture, we visualize the second decoder layer. Our proposed regularization helps the network concentrate better on regions of interest.}
    \vspace{-4ex}
    \label{fig:qualitative_comparison}
\end{figure}

We qualitatively investigate our proposed method's regularization effects through visualizations of the internal activations, and specifically, we visualize the second decoder layer for both trained U-Net and DenseDepth, as shown in \Cref{fig:qualitative_comparison}. In general, our proposed regularization helps the network concentrate on regions of interest at the early decoding stage.

For the U-Net without regularization, the internal activation of the trained network shown in \Cref{fig:unet_comparison} concentrates evenly on nearly everywhere in the image. This might help the network gather local information but it's not necessary for dense estimation problems. This is due to the fact that dense prediction problems tend to have region blocks whose internal gradients are minimal. Thus, it's better for the network to concentrate on regions than on pixels during the early decoding stage, and the fine-grain information can be later accumulated through skip connections from early encoding stages. In contrast, our regularized model learns to focus on high-level regions. Our regularization also aids in reducing artifacts after unpooling and deconvolution seen in activations.

For DenseDepth, the trained network without regularization already concentrates on regions, as shown in \Cref{fig:densedepth_comparison}. Our proposed regularization further strengthens the level of concentration. The foreground lamps, bushes and trees tend to have lower values in our proposed method, and building regions now tend to have higher values.

In both models, we see that topological regularization aids the decoder in focusing on important regions of the image.  This also provides a high level of interpretability into how successful networks are making inferences while abstracting away detail about the particular activations.

\subsection{Quantitative comparison}
\begin{figure}[tp]
    \centering
    \includegraphics[width=\linewidth]{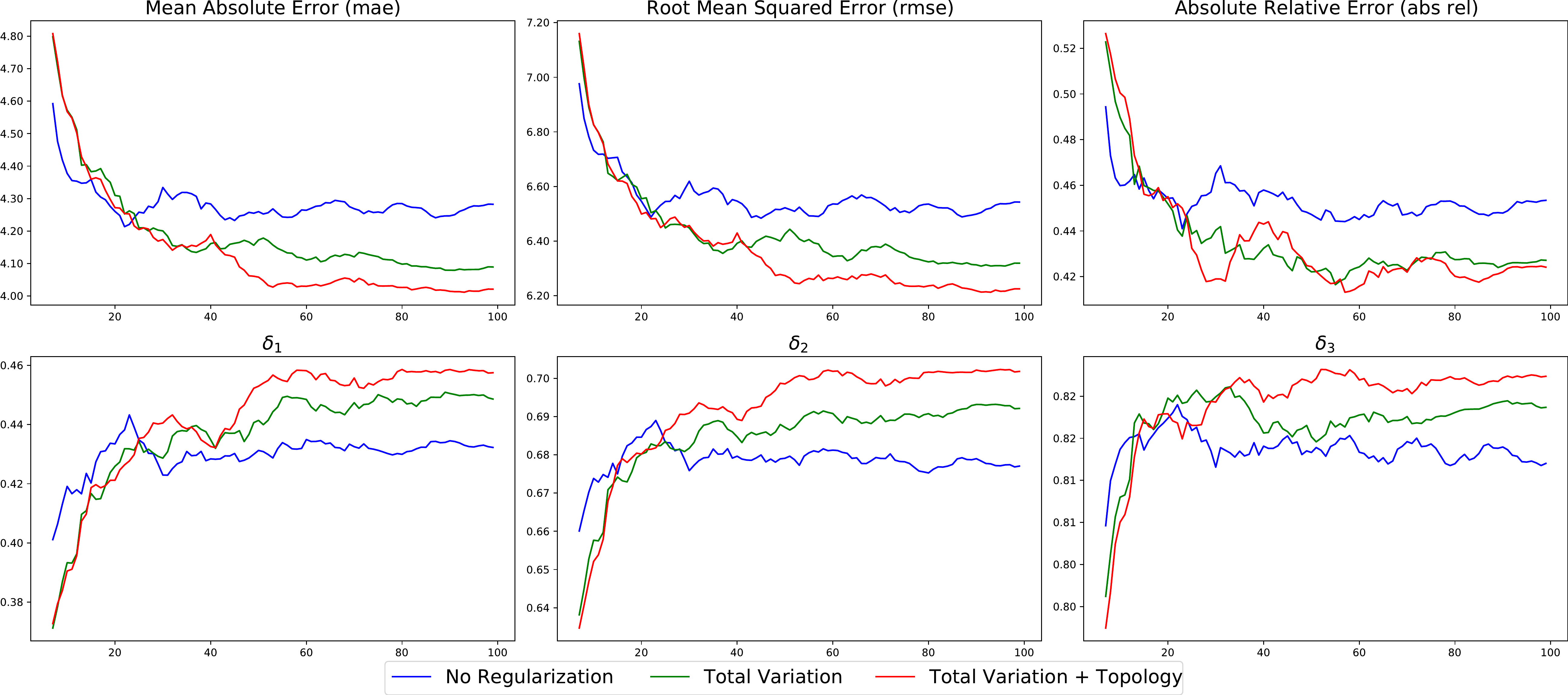}
    \caption{\textbf{Improvements on U-Net Convergence.} Blue lines (without regularization) demonstrates some degree of overfitting as loss increases and accuracy decreases after 20 epochs. Green lines (with total variation regularization) alleviates overfitting a bit, and Red lines (with both proposed regularizations) further overcomes overfitting and converges to a better optimum.}
    \vspace{-3ex}
    \label{fig:convergence}
\end{figure}

\noindent \textbf{Benchmark Metrics.} We quantitatively compare the performance of our regularization on both U-Net and DenseDepth architectures. The accuracy metrics are defined as the following, 
\begin{enumerate}[i.,topsep=2pt,itemsep=0.5ex]
\small 
    \item Mean Absolute Error (mae): $\frac{1}{h w} \sum_{i,j} |y_{i,j} - \hat{y}_{i,j}|$
    \item Root Mean Squared Error (rmse): $\sqrt{\frac{1}{h w} \sum_{i,j} (y_{i,j} - \hat{y}_{i,j})^2}$
    \item Absolute Relative Error (abs rel): $\frac{1}{h w} \sum_{i,j} \frac{|y_{i,j} - \hat{y}_{i,j}|}{\hat{y}_{i,j}}$
    \item mae $\log_{10}$: $\frac{1}{h w} \sum_{i,j} |\log_{10} y_{i,j} - \log_{10} \hat{y}_{i,j}|$
    \item rmse $\log_{10}$: $\sqrt{\frac{1}{h w} \sum_{i,j} (\log_{10} y_{i,j} - \log_{10} \hat{y}_{i,j})^2}$
    \item Threshold accuracy ($\delta_k$): Percentage of pixels such that $\max \left(\frac{y_{i,j}}{\hat{y}_{i,j}}, \frac{\hat{y}_{i,j}}{y_{i,j}} \right) = \delta_k < 1.25^k$. We specifically care about when $k = 1,2$ and 3. 
\end{enumerate}

\noindent \textbf{Convergence Improvement.} Our proposed method is also advantageous in speed up the converge. \Cref{fig:convergence} plots the U-Net model's convergence curves of both validation losses and validation threshold accuracies. The vanilla model without regularization demonstrates some degree of overfitting as the validation losses start to increase and accuracies decrease after 20 epochs. Adding a total variation regularization helps alleviate overfitting but also shows slight decrease in $\delta_3$ accuracy after 40 epochs. Enforcing an additional topological regularization further alleviates the overfitting pattern and converges to a better optimum. 

\noindent \textbf{Performance Improvement.} We benchmark our proposed method on the entire DIODE dataset of both indoor and outdoor scenes. The benchmark results are listed for both models, U-Net  and DenseDepth in \Cref{tab:performace}. Generally, our regularized model can achieve better performance on nearly every metric, compared with the unregularized model. 

For U-Net, we further study the effects of different levels of regularization. By simply adding a total variation regularization, we can already reduce the losses and increase the threshold accuracy. As we subsequently add the proposed topological regularization, the scores improve further. In terms of threshold accuracy, we can improve about 2.3\% in $\delta_1$, 2.5\% in $\delta_2$ and 1.0\% in $\delta_3$. 

We further develop the same regularization a recent state-of-the-art network, DenseDepth, and our proposed method can also help improve benchmark scores. In terms of threshold accuracy, we can improve about 1.1\% in $\delta_1$, 1.0\% in $\delta_2$ and 1.2\% in $\delta_3$. For validation losses, the only metric our proposed method performs worse on is the absolute relative error. This may due to the fact that our regularization is based on super-level set filtration, which will naturally focus on high-value regions, and in depth estimation problems, these regions would be those at the background. In this sense, our proposed method may perform slightly worse on foreground objects than the baseline model without regularization. 

%% file: 06conclusion.tex
\vspace{-2ex}
\section{Conclusion}
\label{sec:conclusion}

We have shown how super-level set topology plays an important role in semantic segmentation and monocular depth estimation problems.  This offers a high level of interpretability into the internal working of neural networks trained to solve these problems, and we show this can be used to regularize training for faster convergence and increased accuracy.  We anticipate these insights will be applicable to other dense prediction problems, and our topological regularization techniques may be adapted to a variety of architectures.

Avenues for future work may include extending topological regularizations to videos where consecutive frames are likely to share similar topological structures. Although this work demonstrates the existence of topological structures in learned convolutional neural networks and shows a way of regularizing internal activations, whether the phenomenon and regularization are also valid for the paradim of vision transformers (ViT) \cite{Dosovitskiy2021AnII} is unknown and of future interest. 




We believe that describing the behavior of internal activations of neural networks using topology has great potential for explaining how neural networks operate in practice.  Persistent homology has the advantage of abstracting away details about the individual weights and activations in the network while retaining important geometric information -- in our case the prominence of local maxima.  Regularization is one method of leveraging insights from topology to improve neural networks which we have applied to dense prediction.  There are also efforts to use topology to initialize weights \cite{rickardCNN2019} and inform decisions in network architecture \cite{naitzat2020} which may also be applicable to dense predictions.

\iftrue   
\vspace{-2ex}
\section*{Acknowledgements}
DF would like to thank Michael Maire for discussions and feedback. 
BN was supported by the Defense Advanced Research Projects Agency (DARPA) under Agreement No. HR00112190040.
We thank the Department of Computer Science, University of Chicago and Toyota Technological Institute at Chicago for providing cluster resources. 
\fi